\documentclass[lettersize,journal]{IEEEtran}
\usepackage{amsmath,amsfonts}
\usepackage{algorithmic}
\usepackage{algorithm}
\usepackage{array}
\usepackage[caption=false,font=normalsize,labelfont=sf,textfont=sf]{subfig}
\usepackage{textcomp}
\usepackage{stfloats}
\usepackage{url}
\usepackage{verbatim}
\usepackage{graphicx}
\usepackage{cite}
\usepackage{threeparttable}
\usepackage{adjustbox}
\usepackage{hyperref}

\usepackage{caption}
\usepackage[most]{tcolorbox}

\newcommand{\eg}{\text{e.g.}}
\newcommand{\ie}{\text{i.e.}}

\begin{document}

\title{Unveiling Effective In-Context Configurations for Image Captioning: An External \& Internal Analysis}

\author{Li Li, 
        Yongliang Wu, Jingze Zhu, Jiawei Peng, 
        Jianfei Cai,~\IEEEmembership{Fellow,~IEEE,} 
        and Xu Yang
\thanks{Li Li, Yongliang Wu, Jingze Zhu, and Jiawei Peng are with the School of Computer Science \& Engineering, Key Lab of New Generation Artificial Intelligence Technology \& Its Interdisciplinary Applications (Ministry of Education), Southeast University, China}
\thanks{Jianfei Cai is currently a Professor and serving as the Head for the Data Science \& AI Department at Faculty of IT, Monash University, Australia.}
\thanks{Xu Yang is currently a Professor at School of Computer Science \& Engineering, Southeast University, China.}
}

\markboth{Journal of \LaTeX\ Class Files,~Vol.~14, No.~8, August~2021}%
{Shell \MakeLowercase{\textit{et al.}}: A Sample Article Using IEEEtran.cls for IEEE Journals}


\maketitle

\begin{abstract}

The evolution of large models has witnessed the emergence of In-Context Learning (ICL) capabilities. In Natural Language Processing (NLP), numerous studies have demonstrated the effectiveness of ICL. Inspired by the success of Large Language Models (LLMs), researchers have developed Large Multimodal Models (LMMs) with ICL capabilities. However, explorations of demonstration configuration for multimodal ICL remain preliminary. Additionally, the controllability of In-Context Examples (ICEs) provides an efficient and cost-effective means to observe and analyze the inference characteristics of LMMs under varying inputs.
This paper conducts a comprehensive external and internal investigation of multimodal in-context learning on the image captioning task. 
Externally, we explore demonstration configuration strategies through three dimensions: shot number, image retrieval, and caption assignment. We employ multiple metrics to systematically and thoroughly evaluate and summarize key findings.
Internally, we analyze typical LMM attention characteristics and develop attention-based metrics to quantify model behaviors. 
We also conduct auxiliary experiments to explore the feasibility of attention-driven model acceleration and compression. We further compare performance variations between LMMs with identical model design and pretraining strategies and explain the differences from the angles of pre-training data features.
Our study reveals both how ICEs configuration strategies impact model performance through external experiments and characteristic typical patterns through internal inspection, providing dual perspectives for understanding multimodal ICL in LMMs.
Our method of combining external and internal analysis to investigate large models, along with our newly proposed metrics, can be applied to broader research areas.

\end{abstract}


\section{Introduction}

\IEEEPARstart{I}{n} recent years, the emergence of Large Language Models (LLMs)~\cite{brown2020gpt3, devlin2018bert, radford2019language, touvron2023llama} has significantly transformed the field of Natural Language Processing (NLP). Prompt engineering~\cite{liu2023promptsurvey, white2023prompt, brown2020gpt3}, a key innovation, allows various NLP tasks to be reformulated as language modeling tasks, leveraging versatility of LLMs without requiring extensive task-specific training. This has led to the development of specialized techniques like In-Context Learning (ICL)~\cite{brown2020gpt3, su2022nlpicl, gao2020nlpicl,jiang2020nlpicl} and Chain-of-Thought (CoT) prompting~\cite{wei2022cot}. ICL, in particular, has proven effective for complex tasks that are difficult to handle with simple prompts, by providing few-shot examples to clarify task requirements.
Researches~\cite{brown2020gpt3, rubin2021nlpretrieve,zhang2022nlpretrieve,wu2022nlpretrieve,gonen2022nlpretrieve} have shown that LLMs are sensitive to the configuration of In-Context Examples (ICEs), such as their selection~\cite{liu2021selection} and order~\cite{lu2021order}. This has inspired numerous studies~\cite{tanwar2023nlpicl, su2022nlpicl, kim2022self, gao2020nlpicl} focused on developing strategies to optimize ICEs configurations for performance improvement. ICL also provides a controlled framework for investigating the internal properties of LLMs by allowing researchers to manipulate the composition of ICEs~\cite{min2022rethinking,tr&tl}.

Following the success of LLMs in NLP, researchers have extended these capabilities to Large Multimodal Models (LMMs)~\cite{alayrac2022flamingo, awadalla2023openflamingo,radford2021clip,zhu2023minigpt4,liu2023llava, 9843903, YangZWLXJ21, laurenccon2024obelics}. These models are typically trained in two steps: multimodal pre-training to align visual and textual modalities, and Visual Instruction Tuning to adapt to new tasks using vision-conditioned instruction samples. 
This enables LMMs to perform vision-language tasks like Image Captioning~\cite{lin2014mscoco} and Vision Question Answering~\cite{goyal2017vqav2,marino2019okvqa}. 
However, LMMs initially showed limited ICL abilities. To address this, models like Flamingo~\cite{alayrac2022flamingo} and its open-source version OpenFlamingo~\cite{awadalla2023openflamingo} were developed. They introduced gated cross-attention dense blocks to enhance vision-language alignment and were trained in few-shot settings to optimize ICL performance. Subsequently, IDEFICS~\cite{laurenccon2023IDEFICS}, based on the Flamingo framework and trained on the web-collected OBELICS dataset, demonstrated significant improvements in multimodal ICL capabilities.



Though LMMs have shown ICL capabilities, most studies use simple ICEs configuration methods like random sampling or basic similarity-based retrieval~\cite{liu2021selection}. This approach overlooks the sensitivity of LMMs to ICEs configurations, limiting the realization of their full ICL potential.
Moreover, configuring ICEs for vision-language tasks is more complex than for single-domain NLP tasks. It requires considering the dual-modal synergy between vision and language, rather than just the characteristics of individual modal elements. 
For instance, although using similar retrieval usually improves performance in NLP, our results indicate that similar ICE images with low-quality captions can negatively impact outcomes. Conversely, dissimilar images can partially counteract poor text quality.
This highlights the importance of studying ICEs configurations in LMMs, emphasizing the need to account for the integrated characteristics of vision-language elements.



Meanwhile, large model reasoning often remains a ``black box" due to end-to-end pre-training, nonlinear neural transformations, and complex high-dimensional features. However, in-context learning provides a unique way to observe and analyze model reasoning. By manipulating contextual examples, researchers can study the responses of the model to different inputs. This sensitivity to context and the controllability of ICL configurations make it suitable for conducting controlled experiments and mechanistic analyses of model reasoning.
However, systematic and comprehensive investigations into model reasoning remain limited.
Moreover, the reasoning differences in large models often stem from variations in their pre-training datasets, \eg, scale~\cite{kaplan2020scalinglaw} and data characteristics\cite{zhao2024pretraindata, cheng2025pretraindata}. However, studying these effects with controlled variables by training and comparing multiple large models from scratch is too costly. Our approach, leveraging ICL configurations, allows for a more economical and efficient comparison of LMMs trained on different datasets.


Briefly, this study examines how different configurations affect LMM in-context learning and the factors driving performance variations. 
We adopt a two-pronged approach that mirrors the scientific method used by researchers to uncover the mysteries of the objective world, \ie,  observing the effects of varying configurations from an external perspective and analyzing the inner dynamics of a system from an internal perspective. 
Externally, similar to how scientists conduct controlled experiments, we vary in-context configurations to observe their impact on model performance. 
Internally, akin to a biologist studying an organism anatomy, we analyze LMM attention weights to understand vision-language integration. 
By combining these approaches, we comprehensively analyze how in-context configurations affect model reasoning and reveal LMM advantages and disadvantages, offering insights for refining current models.
While our research focuses on multimodal in-context learning, the integrated approach of examining both external and internal perspectives can be applied to a broader range of exploration fields.

In this research, we select Image Captioning~\cite{lin2014mscoco} as our case study to explore in-context configurations for LMMs. This choice is justified by several key reasons. Firstly, like many NLP tasks---from text summarization~\cite{el2021textsum} to question answering~\cite{allam2012questionans}---that can be reformulated as Language Modeling (LM) tasks, Image Captioning can also be seen as a specific LM task where the goal is to generate text based on the visual context from the source image. Secondly, among various vision-language tasks, Image Captioning has a relatively straightforward format, with each in-context example consisting of one image and one text. This simplicity facilitates the development of ICEs configuration strategies and the analysis of information flow between visual and linguistic modalities. Thirdly, Image Captioning offers versatile evaluation methods\cite{vedantam2015cider,hessel2021clipscore,rohrbach2018chair}. Metrics such as CIDEr and CLIPScore assess the similarity between generated text and ground truth text and image, while metrics like CHs and CHi detect textual hallucinations. After selecting Image Captioning as our case study, we design various methods to analyze the ICL capabilities of LMMs from both external and internal perspectives.
We conduct experiments on two LMMs, OpenFlamingo and IDEFICS. Both are based on the Flamingo framework and have similar architectures, but they use different pre-training datasets. Consequently, they exhibit different performance characteristics in our experiments.

Externally, in our exploration of demonstration configuration strategies for LMMs, we focus on three key aspects: the ICE image retrieval, the post-retrieval caption assignment, and the ICEs number (\ie, shots). 
These strategies are evaluated using multiple metrics that assess the text quality, visual accuracy, hallucination levels, plagiarism rates, and the utilization of query visual and demonstration text information in the generated captions.
Through these experiments, we have derived six key findings, including increasing shots enhances linguistic coherence but harms visual-text alignment, pre-training corpus bias causes multimodal disparity, ICE quality sensitivity triggers multi-metric conflicts, ICE linguistic style affects visual fidelity, and similarity-based retrieval inflates CIDEr scores via shortcuts while amplifying ICE caption influence.
Internally, we examine the attention maps and design attention-based metrics to measure attention flow within demonstration sequences. Based on initial observations, we identify three key internal characteristics of LMMs: Anchor Token, Emergent Attention Window, and Short-Cut of In-Context Caption. 
To quantify these phenomena, we propose three novel attention-based metrics.These metrics measure the extent of anchor token and emergent attention window patterns and aid in explaining short-cuts and hallucinations, crucial for analyzing experimental results.
Given the sparse attention distribution during ICL of LMMs, we also conduct preliminary lightweight acceleration experiments. 

In summary, this paper makes the following contributions:
\begin{itemize}
\item[$\bullet$] 
We conduct a comprehensive analysis of multimodal in-context learning through thorough external and internal investigations on two large multimodal models.
\item[$\bullet$] 
We derive six significant conclusions in external experiments and identified three typical characteristics in internal analysis.
\item[$\bullet$] 
We design useful external metrics for text generation and attention-based internal metrics with potential applications in other domains.
Our integrated approach of examining both external and internal perspectives supports broader exploration across fields.
\item[$\bullet$] 
We compare the performance of two architecturally similar LMMs under identical settings and explain the differences from the angles of pre-training data features. This offers guidance for future LMM design and training and provides a fast and cost-effective way to explore the impact of pre-training data on model performance.
\end{itemize}

This paper is an extension of our preliminary work~\cite{wylnips} with
significant modifications including
\begin{itemize} \itemsep0pt
	\item[$\bullet$] We expand the experimental scope to include more LMMs, which enhances the generalizability of our conclusions.
	\item[$\bullet$] A new section on internal attention analysis has been added, which explores unique attention patterns in multimodal ICL, develops novel attention-based metrics, and performs preliminary inference acceleration experiments.
	\item[$\bullet$] We provide a more comprehensive analysis of the experimental results related to the configurations. By introducing new metrics and combining them with attention distribution analysis and model pre-training differences, we offer deeper insights into the experimental conclusions, assisting in the training and exploration of future LMMs.
\end{itemize}

\section{Related Works}
\subsection{Large Language Model Based Multimodal Models}

After observing that LLMs exhibit zero-shot learning abilities through appropriate pre-training~\cite{devlin2018bert,openai2023gpt,radford2019gpt2,brown2020gpt3}, similar efforts have been made to extend this zero-shot capability to the multimodal domain. CLIP~\cite{radford2021clip} is an early example, trained via contrasting images and captions to achieve zero-shot learning, though its ability is limited to coarse-grained tasks~\cite{li2023scaling}. To achieve more fine-grained zero-shot learning, researchers align a vision encoder with a well-trained LLM, creating more robust LMMs~\cite{lu2019vilbert,chen2020uniter,li2020oscar, Li2022mPLUGEA,zhu2023minigpt4,liu2023llava,li2022blip,yu2022coca}.


As LLMs demonstrate impressive few-shot or in-context learning abilities~\cite{brown2020language}, multimodal researchers aim to empower LMMs with similar capabilities. Unlike LLMs, whose training data naturally contain implicit input-output pairs~\cite{wang2018glue}, multimodal data lack such pairs, requiring artificial dataset construction. Additionally, although there are various tasks in NLP, almost all of the input-output pairs can be uniformly represented in texts and thus LLMs can use more unified network structure like a Transformer~\cite{vaswani2017transformer} only including self-attention layers to deal with the data. However, images and texts belong to different modalities, increasing the design complexity for LMMs to handle interleaved vision-language data. Early LMMs like MiniGPT-4~\cite{zhu2023minigpt4} and BLIP-2~\cite{li2023blip2} struggled with these issues, limiting their ICL capabilities.


To address these challenges, Flamingo~\cite{alayrac2022flamingo}, with its open-source variant OpenFlamingo~\cite{awadalla2023openflamingo}, was developed, constructing interleaved vision-language data and applying dense cross-attention gates in the network, significantly improving ICL performance in multimodal tasks. Subsequently, IDEFICS~\cite{laurenccon2023IDEFICS}, another Flamingo architecture model, has refined a vision-language dataset, OBELICS, to further enhance ICL abilities. 
While there are other models with ICL capabilities, some are not open-sourced, and others use benchmark datasets like MSCOCO~\cite{lin2014mscoco} for pre-training, thus inefficient in validating the generalization ability. Considering these factors, we choose OpenFlamingo and IDEFICS for our experiments.

\begin{figure*}[!ht]
\vspace{-15pt}
  \centering
  \includegraphics[width=1.0\linewidth]{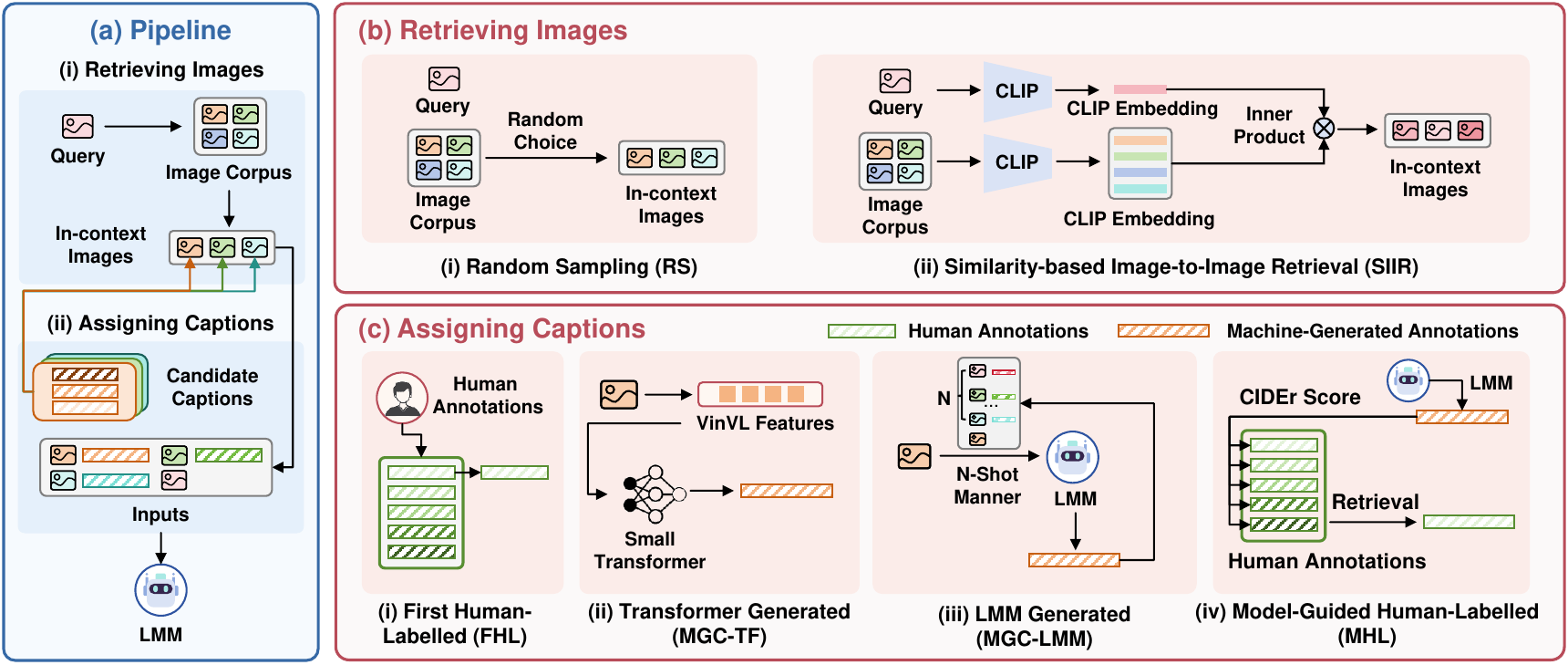}
  \vspace{-15pt}
  \caption{\textbf{External Configuration Strategies.} (a) The overall pipeline: it first retrieves in-context images from an image corpus, then constructs demonstration sequences by assigning a corresponding caption to each in-context image. (b) Two image-retrieving methods: Random Sampling (RS) and Similarity-based Image-to-Image Retrieval (SIIR). (c) Four caption-assigning methods: using human-labelled ground-truth annotations, machine-generated ones (from a transformer or LMM itself), and selecting the most similar ground-truth annotation with machine-generated ones.
}
\vspace{-10pt}
  \label{fig:external_pipeline}
\end{figure*}

\subsection{Configuring ICL}

Extensive studies in LLMs ~\cite{gao2020nlpicl,jiang2020nlpicl,su2022nlpicl,levy2022diverse,lu2021order, Qin2023InContextLW} have demonstrated that ICL performance is critically influenced by configuration factors including demonstration selection strategies~\cite{liu2022makes, an2023context,zhang2022active}, label composition~\cite{lyu2022z}, and demonstration ordering~\cite{lu2022fantastically,liu2022makes}. Current optimization approaches can be categorized into two paradigms:
heuristic-based and learning-based.
Heuristic-based methods primarily leverage semantic metrics to curate demonstrations. Early work focuses on retrieving examples with maximal semantic similarity to the query~\cite{liu2022makes}, while subsequent extensions incorporate diversity-aware selection criteria that balance relevance with category coverage~\cite{levy2022diverse}. These strategies effectively enhance ICL stability through controlled demonstration ensembles.
Learning-based approaches adopt data-driven frameworks to optimize ICEs configurations. A prominent direction involves training dedicated example encoders on annotated data to capture semantic features critical for context-aware selection~\cite{wu2022self}. Alternative formulations frame the configuration process as sequential decision-making tasks, enabling end-to-end optimization of demonstration impacts.

Given that the work mentioned above focuses on the NLP area within a single modality, some studies~\cite{wylnips, xu2024lmmicl,luo2024lmmicl,li2024configure, peng2023icd, baldassini2024lmmicl} have begun to explore the issue of ICL configurations for LMMs. 
However, most existing studies have only explored simple and intuitive demonstration configuration strategies, falling short of a comprehensive investigation into multimodal ICL. There is also a lack of explanations for why certain strategies are effective or not.
Meanwhile, single-modal optimization methods struggle with cross-modal heterogeneity in multimodal tasks. 
Mismatches in image-text feature spaces can cause semantic conflicts~\cite{radford2021learning}, and imbalances of modal parameters favoring the language module lead to ignored visual clues~\cite{li2024configure, lyu2022z}. Existing strategies fail to detect contradictions from cross-modal information mismatches that cause reasoning deviations.

\subsection{Analyzing ICL}
Research~\cite{xie2021explanation} has proposed that the inference process of ICL can be viewed as Bayesian inference on a shared concept, where a concept is considered a latent variable containing textual statistical information. One study~\cite{min2022rethinking} conducts experiments where samples and ground truths in ICEs are intentionally mismatched, and find that this only has a minor impact on performance. They argue that the performance gains of ICL primarily stem from the independent regularization of input and label spaces, as well as from a correct and consistent demonstration format. Additionally, research~\cite{shi2023larger} further indicates that when LLMs reach a sufficient scale, they exhibit an emergent capability to comprehend input-label mappings, regardless of whether the labels are inverted or semantically-unrelated. 
From the internal perspective of models, the analysis of attention distribution reveals that example labels act as anchors during reasoning, guiding semantic information aggregation in deeper layers and influencing final predictions~\cite{wang2023label}.
Multimodal in-context learning is more complex than its single-model counterpart. 
In this work, we consider analyzing the inference of LMM ICL from both internal and external perspectives, and focus on the differences between the two modalities.

\section{Approach}

\subsection{ICL Formulation in LMM}
Formally, given a pretrained Large Multimodal Model \(M\), a task \(T\), and a query sample image \(\hat{I}\), the aim of in-context learning is to select \(n\) task-related ICEs \(E = \{(I_1, c_1), (I_2, c_2), \ldots, (I_n, c_n)\}\) from demonstration sets \(D\). Here, \(I_i\) denotes the image for the \(i\)-th example in the context and \(c_i\) denotes the corresponding caption. The output \(c\) for the image query \(\hat{I}\),  is then generated by utilizing the context \(E\) as condition. This process can be formally represented as:

\[
c = M(\hat{I} | E).
\]

\subsection{External Configuration Strategies\label{configuration method}}
Considering Image Captioning is a vision-language task, each in-context example consists of two parts: an image and a caption. Therefore, in external configuration strategies, we have set different strategies for selecting both the image and the caption to observe the impact of different combinations on the final results.
As shown in Figure~\ref{fig:external_pipeline} (a), our demonstration configuration strategy has two steps. First, we retrieve n-shot ICE images from the image corpus using different methods. Then, we choose a caption from different candidate annotations for each image. Combined with the query image, these form the input demonstration sequence.

\subsubsection{Retrieving Images}
In many previous studies in NLP \cite{liu2021selection,kim2022self, wylnips,li2024configure}, it has been found that in ICL, the similarity between ICE and the query can affect the final performance. Therefore, when retrieving images as in-context images, our main consideration is to control the similarity between the images and the query to observe the impact of similarity on the final generated captions. 
When selecting captions, we have set two levels of similarity as shown in Figure~\ref{fig:external_pipeline} (b): randomly selecting images from the supporting set, which is called Random Sampling (RS), and selecting images most similar to the query image, which is called Similarity-based Image-to-Image Retrieval (SIIR).
When calculating image similarity, we use the CLIP \cite{radford2021clip} vision encoder to embed the images and then calculate the cosine similarity between the embeddings as the similarity score.

\subsubsection{Assigning captions}
After retrieving in-context images, we should assign a caption to each retrieved image. When assigning a caption to each in-context image, our main consideration is to control the quality of caption to observe the impact of varying quality on the final outcome. 
In controlling the quality of caption, we use four different types of captions in Figure~\ref{fig:external_pipeline} (c). 
The first is the most basic and common setting, which directly uses the first ground-truth human-annotated caption, akin to randomly selecting, termed First Human-Labelled (FHL). These captions are of high quality due to their human annotation. 
The second type is machine-generated annotations, which includes two subtypes: captions generated by a small Transformer, called Machine Generated Caption-Transformer (MGC-TF), and captions generated by the LMM itself, named Machine Generated Caption-LMM (MGC-LMM).  These captions, generated by models, typically have relatively lower quality compared to human-annotated ones. However, they may contain the preferences of generative models, which could lead to unexpected effects.
The last type is Model-Guided Human-Labelled (MHL), which 
includes the machine-generated annotation as reference points to select the human-annotated caption that is most similar to this model-generated caption, where similarity can be calculated using CIDEr. 
Furthermore, to evaluate the impact of text quality in demonstrations on inference, we test captions of varying quality. 
For captions generated by the small Transformer (MGC-TF), we use CIDEr as the training objective. These include captions from models that have not fully converged during training (low quality) and those from models that have reached convergence (high quality).
For the captions generated by the LMM itself (MGC-LMM), we first use the 0-shot/32-shot ICL method to generate captions, with 0-shot generation representing low quality and 32-shot generation representing high quality. These generated captions are then utilized as the in-context captions for subsequent evaluations.

\subsection{Internal Analysis Strategies\label{internal}}
Since the proposal of Transformer~\cite{vaswani2017transformer} and BERT~\cite{devlin2018bert}, the academic community has increasingly utilized attention maps to analyze the internal characteristics of language models~\cite{kovaleva2019llmanchor}. This approach has been extended in numerous subsequent studies to investigate the internal features of LLMs~\cite{wang2023llmanchor,huang2024lmmanchor,liao2024anchorllm,zhu2024anchor&window}. Here, we adopt a similar methodology to scrutinize the internal characteristics of LMMs during the execution of in-context learning.
In this section, specifically, we concentrate on three internal characteristics which are termed as \textbf{Anchor Token}, \textbf{Emergent Attention Window} and \textbf{Short-Cut of In-Context Caption}.
For each internal characteristic, we initially conduct a qualitative analysis by examining the attention maps, followed by a quantitative analysis utilizing attention scores. 


\begin{figure*}[!ht]
\vspace{-5pt}
  \centering
  \includegraphics[width=1.0\linewidth]{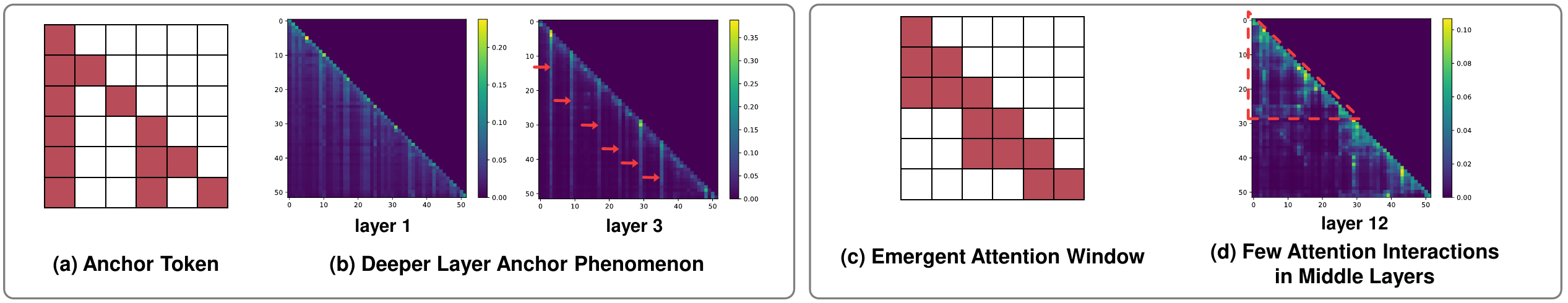}
  \vspace{-15pt}
  \caption{\textbf{Two typical internal attention patterns.} (a) Anchor Token. (c) Emergent Attention Window. (b) and (d) show a visualization example of these two patterns (4-shot, IDEFICS). For greater clarity, we only show the attention maps of the first two ICEs with the initial token dropped. The red arrows in (b) point to the anchor tokens, which are the image token [\textless image\textgreater], the punctuation marks [.] at the end of each ICE, and the delimiters [\textless endofchunk\textgreater] between ICEs. The red dashed triangle in (d) contains an ICE. It is evident that there are more attention interactions within the box, while the attention weights between the two ICEs are very low, forming an emergent attention window.
}
\vspace{-10pt}
  \label{fig:attention map}
\end{figure*}

\subsubsection{Anchor Token}

In some studies on LLMs \cite{liao2024anchorllm,wang2023llmanchor,kovaleva2019llmanchor} and LMMs \cite{huang2024lmmanchor,chen2024imagetokenprune}, researchers have observed the presence of anchor tokens in the attention maps of models, as shown in Figure~\ref{fig:attention map} (a). These tokens act as ``anchors'', aggregating information from other tokens in shallow layers and then being predominantly attended to by other tokens in deeper layers. Anchor tokens have been found to exist in both LLMs and LMMs, despite differences in architecture and input, though varying parts of the input sequence may serve as anchor tokens. This raises the question of whether anchor tokens exist in the LMM ICL case and, if so, which tokens in the input sequence play such roles.
%


To identify these anchor tokens, we visualize the attention maps of OpenFlamingo for a given case, as shown in Figure~\ref{fig:attention map} (b). We observe that in the shallow layer (Layer 1), attention interactions are relatively evenly distributed, but in the deeper layer (Layer 3), certain columns appear brighter, indicating increased attention from other tokens. These columns correspond to the initial token [\textless BOS\textgreater], the image token [\textless image\textgreater], punctuation marks [.] at the end of each ICE, and delimiters [\textless endofchunk\textgreater] between ICEs.

\begin{figure*}[!ht]
\vspace{-10pt}
  \centering
  \includegraphics[width=1\linewidth]{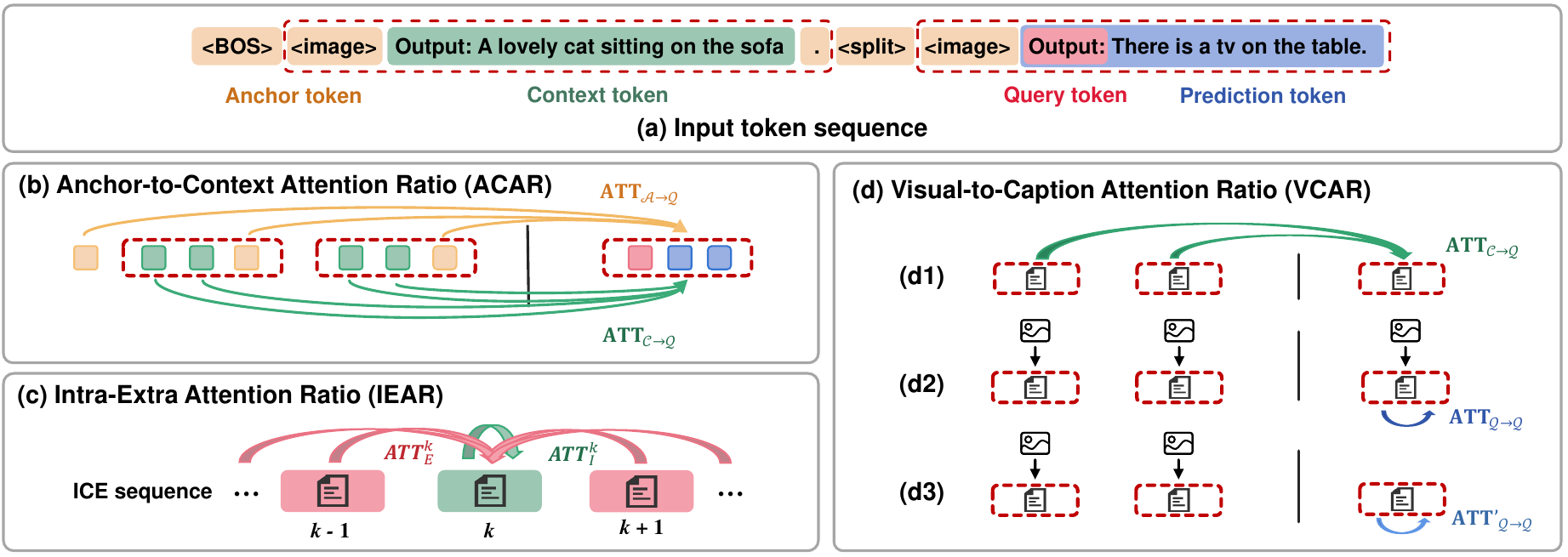}
  \vspace{-15pt}
  \caption{\textbf{Attention-based metrics.} (a) The input token sequence of LMM is divided into different components. 
  (b) ACAR calculation schematic. 
(c) IEAR calculation schematic. 
(d) VCAR calculation schematic. 
}
\vspace{-10pt}
  \label{fig:attention-based metric}
\end{figure*}

To further validate whether these tokens are anchor tokens and assess whether the aggregation pattern exists, we design an attention-based metric: Anchor-to-Context Attention Ratio (ACAR), to quantitatively analyze the attention patterns at each layer. 
Specifically, we categorize input tokens into three classes as shown in Figure \ref{fig:attention-based metric} (a): 
1) Anchor tokens ($\mathcal{A}$): Including the initial token, image token, period marks at the end of each ICE, and delimiters between ICEs.
2) Query tokens ($\mathcal{Q}$): The last two tokens of the input, generally indicating the agent to begin prediction.
3) Context tokens ($\mathcal{C}$): All tokens within the ICE, excluding anchors.
We compute the attention values from the anchor to the query tokens: \textbf{ATT$_{\mathcal{A} \longrightarrow \mathcal{Q}}$} and the values from the context to the query tokens: \textbf{ATT$_{\mathcal{C} \longrightarrow \mathcal{Q}}$} as Figure \ref{fig:attention-based metric} (b) shows. 
ACAR is calculated as the ratio of two attention values:
\begin{equation}
     \text{ACAR} = \frac{ \textbf{\text{ATT}} _{\mathcal{A} \longrightarrow \mathcal{Q}} }{ \textbf{\text{ATT}} _{\mathcal{C} \longrightarrow \mathcal{Q}} },
\end{equation}
\begin{equation}
\textbf{ATT} _{\mathcal{A} \longrightarrow \mathcal{Q}} = \frac{1}{|S_{AQ}|} {\textstyle \sum_{(i,j)\in S_{AQ}  }^{}A_{l}(i,j) } ,
\end{equation}
\begin{equation}
 S_{AQ} = \left \{ (i,j): i \in \mathcal{A}, j \in \mathcal{Q} \right \} ,
\end{equation}
\begin{equation}
\textbf{ATT} _{\mathcal{C} \longrightarrow \mathcal{Q}} = \frac{1}{|S_{CQ}|}{\textstyle \sum_{(i,j)\in S_{CQ}  }^{}A_{l}(i,j) } ,
\end{equation}
\begin{equation}
S_{CQ} = \{ (i,j) \mid i \in \mathcal{C} \setminus \mathcal{A}, j \in \mathcal{Q} \}.
\end{equation}
Here, \( A_l(i, j) \) denotes the attention flow from token \( i \) to token \( j \) at layer \( l \), \( \mathcal{A} \) represents the set of anchor tokens, \( \mathcal{Q} \) represents the set of query tokens, and \( \mathcal{C} \) represents the set of all ICE tokens, \(S\) represents the set of token pairs \((i, j)\), where \(i\) is the key token and \(j\) is the query token.

ACAR can assess the allocation of attention between the anchor and the other context during the prediction, \eg, a higher ACAR indicates a greater reliance on the anchor token, suggesting a more pronounced role of the anchor tokens in mediating the propagation of information throughout the network.
The specific experimental results are presented in Section~\ref{internal results}.




\subsubsection{Emergent Attention Window}
When observing the attention maps when implementing ICL in LMM, we notice that in the middle layers, although we do not manually restrict each token in an ICE to attend only to tokens within the same ICE, it can be found that the tokens naturally attend to each other within the same ICE, with minimal interaction between ICEs as shown in Figure~\ref{fig:attention map} (d). We call this phenomenon the Emergent Attention Window.

To further quantitatively validate this phenomenon, we design an attention-based metric, Intra-Extra Attention Ratio (IEAR), to calculate the average ratio of the attention values within an ICE and between ICEs as shown in Figure\ref{fig:attention-based metric} (c): 
\begin{equation}
     \text{IEAR} = \frac{1}{n} \sum_{k=1}^{n} \frac{ \textbf{ATT} _{I}^{k} }{ \textbf{ATT} _{E}^{k} },
\end{equation}
\(\text{where } n \text{ represents the } n\text{-shot, }\)
\begin{equation}
\textbf{ATT} _{I}^{k} = \frac{1}{\left | S_{I}^{k} \right | }  {\textstyle \sum_{(i,j)\in S_{I}^{k}  }^{}A_{l}(i,j) } ,
\end{equation}
\begin{equation}
 S_{I}^{k} = \left \{ (i,j): i \in \mathcal{I}^{k}, j \in \mathcal{I}^{k} \right \} ,
\end{equation}
\begin{equation}
\textbf{ATT} _{E}^{k} = \frac{1}{\left | S_{E}^{k} \right | } {\textstyle \sum_{(i,j)\in S_{E}^{k}  }^{}A_{l}(i,j) }  ,
\end{equation}
\begin{equation}
S_{E}^{k} = \left\{ (i,j) \mid i \in \mathcal{C} \setminus \mathcal{I}^{k}, j \in \mathcal{I}^{k} \right\} ,
\end{equation}
\(\text{where } \mathcal{I}^{k} \text{ represents the token set of the } k\text{-th ICE.}\)
Here, \(\textbf{ATT} _{I}^{k}\)  represents the average attention interaction of all context tokens within the k-th ICE, and \(\textbf{ATT} _{E}^{k}\) represents the average interaction between context tokens of other ICEs and those of the k-th ICE.
A higher IEAR value indicates more within-ICE interaction and less between-ICE interaction.

\subsubsection{Short-Cut of In-context Caption}

When examining captions generated by diverse ICEs configuration strategies, we find that some strategies make LMMs more prone to directly copy in-context captions during prediction. This phenomenon, termed as ``short-cut inference" \cite{wylnips,li2024configure}, is illustrated in Figure~\ref{fig:visualization} (c) (SIIR setting). Here, the LMM does not generate captions based on the query image but instead use the short-cut path between the in-context caption and the predicted caption.

\begin{figure}[t]
  \centering
  \includegraphics[width=1\linewidth]{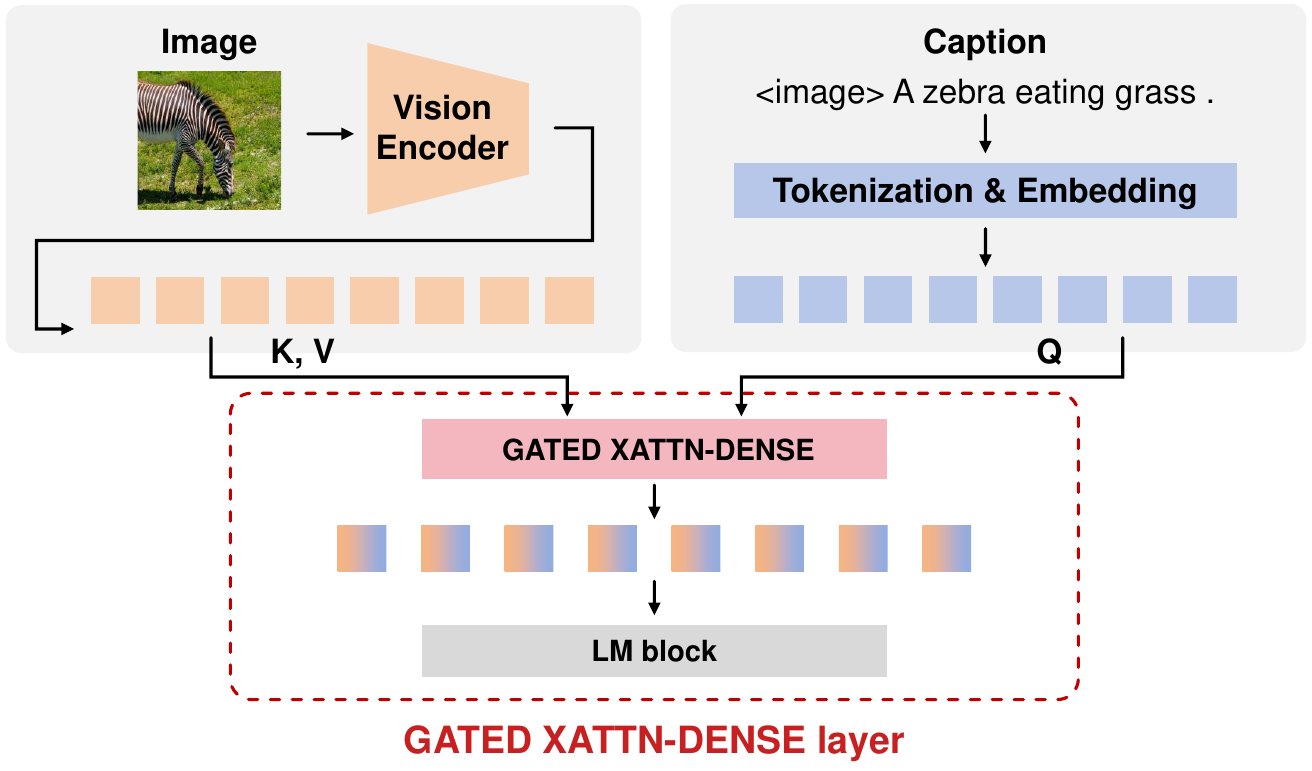}
  \vspace{-15pt}
  \caption{\textbf{The gated cross attention of Flamingo.} Flamingo integrates visual information into text tokens via gated cross-attention layers added to the LLM, thus achieving multimodal capabilities.
}
\vspace{-12pt}
  \label{fig:flamingo_xatt}
\end{figure}


To quantitatively analyze which configuration strategies lead to more short-cut inference, we have developed an attention-based metric called Visual-to-Caption Attention Ratio (VCAR). VCAR measures whether LMMs rely more on visual information from the query image or textual information from in-context captions during generation. 
Ideally, VCAR is calculated by comparing attention values from the query image and in-context captions to generated word tokens. However, in the Flamingo architecture \cite{alayrac2022flamingo}, gated cross-attentions map visual information to paired text, making direct extraction of visual attention values challenging.

To address this, we use an indirect method to estimate visual attention flow by comparing the attention values in two scenarios: one using the query image and one without. The discrepancy between these values reflects the impact of query image on predictions as illustrated in Figure \ref{fig:attention-based metric} (d). 
More specifically, the formula of VCAR can be written as:
\begin{equation}
     \text{VCAR} = \frac{ \textbf{\text{ATT}} _{\mathcal{V_{Q}} \longrightarrow \mathcal{Q}} }{ \textbf{\text{ATT}} _{\mathcal{C} \longrightarrow \mathcal{Q}} },
\end{equation}
\begin{equation}
\textbf{ATT} _{\mathcal{V_{Q}} \longrightarrow \mathcal{Q}} =  \textbf{ATT} _{\mathcal{Q} \longrightarrow \mathcal{Q}} - \textbf{ATT} _{\mathcal{Q} \longrightarrow \mathcal{Q}}^{\prime},
\end{equation}
where \(\textbf{ATT} _{\mathcal{Q} \longrightarrow \mathcal{Q}}\) and \(\textbf{ATT} _{\mathcal{Q} \longrightarrow \mathcal{Q}}^{\prime}\) are the average attention interactions of the query token with itself when reasoning with and without the query image respectively.
\begin{equation}
\textbf{ATT} _{\mathcal{Q} \longrightarrow \mathcal{Q}} = \frac{1}{\left | S_{QQ} \right | }  {\textstyle \sum_{(i,j)\in S_{QQ}  }^{}A_{l}(i,j) } ,
\end{equation}
\begin{equation}
 S_{QQ} = \left \{ (i,j): i \in \mathcal{Q}, j \in \mathcal{Q} \right \} ,
\end{equation}
\begin{equation}
\textbf{ATT} _{\mathcal{C} \longrightarrow \mathcal{Q}} = \frac{1}{\left | S_{CQ} \right | } {\textstyle \sum_{(i,j)\in S_{CQ}  }^{}A_{l}(i,j) }  ,
\end{equation}
\begin{equation}
 S_{CQ} = \left \{ (i,j): i \in \mathcal{C}, j \in \mathcal{Q} \right \}.
\end{equation}
Figure \ref{fig:attention-based metric} (d) shows the calculation process of  \(\textbf{ATT} _{\mathcal{C} \longrightarrow \mathcal{Q}}\), \(\textbf{ATT} _{\mathcal{Q} \longrightarrow \mathcal{Q}}\) and \(\textbf{ATT} _{\mathcal{Q} \longrightarrow \mathcal{Q}}^{\prime}\).


A higher VCAR indicates greater use of visual cues and lower reliance on short-cuts from in-context captions. This metric is essential for understanding how the balance between visual and textual inputs affects the final predictions.
In subsequent experiments, we observe that lower VCAR values, indicating more attention to ICE captions than the query image, are associated with increased hallucination and short-cut behaviors in LMMs. This suggests that excessive focus on ICE captions may lead to short-cuts in prediction. However, while lower VCAR values often correlate with more hallucinations and short-cuts, they do not exhibit a strictly negative correlation. 
This suggests VCAR serves as a partial proxy for diagnosing short-cut behaviors, yet the underlying mechanisms driving hallucinations demonstrate multi-factorial etiology that cannot be fully captured by this singular metric.

\section{Experiments and Analyses}
In Section~\ref{configuration method}, we apply numerous strategies to configure diverse in-context sequence across two different LMMs. However, showcasing all outcomes indiscriminately may lead to organizational chaos. To enhance clarity, we present the pertinent results aligned with each ongoing analysis in appropriate formats, such as tables, line graphs, or histograms. 
\subsection{Datasets and Settings}
\noindent\textbf{Datasets and LMMs.} 
We evaluate the proposed strategies on MSCOCO~\cite{lin2014mscoco} dataset. 
For MSCOCO, we use the Karpathy split~\cite{karpathy2015deep} in the experiments, which contains 113,287/5000/5000 training/validation/test images and each image is associated with 5 human-annotated captions. 
Two state-of-the-art large multimodal models, OpenFlamingo v2-9B~\cite{awadalla2023openflamingo}, and IDEFICS v1-9B~\cite{laurenccon2024obelics} are employed for evaluating. 

\noindent\textbf{Settings.}
For the OpenFlamingo models, we set the temperature to 0.2, while for IDEFICS, the temperature was set to 0.1. Across all tasks, we experimented with shot counts of 4, 8, 16, and 32. All experiments were conducted on a single RTX-3090 GPU using FP16 precision.

\begin{table}[t]
\vspace{0pt}
\centering
\caption{Comparison of OpenFlamingo v2 and IDEFICS v1}
\begin{tabular}{lll}
\hline
\textbf{Model Feature} & \textbf{OpenFlamingo} & \textbf{IDEFICS} \\
\hline
Vision Encoder & CLIP ViT & OpenCLIP ViT \\
Language Encoder & MPT & LLaMA \\
Dataset & LAION, MMC4 & LAION, OBELICS,... \\
Dataset Characteristics & Fewer Tokens & Long Sequences \\
\hline
\\
\end{tabular}
\vspace{-10pt}
\label{OFvsIDE}
\end{table}

\subsection{Preliminaries}
Before presenting the main conclusions of this study, it is essential to first provide some key prerequisite information needed for conclusion analysis.

\subsubsection{Differences Between OpenFlamingo v2 (OF) and IDEFICS v1 (IDE)\label{OFvsIDE}}
OF and IDE share a similar network architecture, each comprising a vision encoder and a language decoder. They both utilize gated cross-attention dense blocks, following Flamingo~\cite{alayrac2022flamingo}, to integrate vision and language components. Their vision encoders are both based on CLIP~\cite{radford2021clip}. Although they employ different language encoders (OF uses MPT \footnote{https://www.databricks.com/blog/mpt-7b} and IDE uses LLaMA \cite{touvron2023llama}), there is no significant performance gap between these two language models.


The performance gap between the two LMMs largely stems from the used training datasets. While both are partially trained on LAION-2B~\cite{schuhmann2022laion}, IDE also uses OBELICS~\cite{laurenccon2024obelics}, whereas OF uses MMC4~\cite{zhu2023mmc4}.
The datasets used during the pre-training phase of these models have a substantial impact on their differential performance, particularly in terms of multimodal data alignment. Specifically, MMC4, used for pre-training OF,  has been filtered from the C4\footnote{https://www.tensorflow.org/datasets/catalog/c4} dataset through CLIP, resulting in higher CLIPScores~\cite{hessel2021clipscore} in our tests. However, the dataset contains a significant number of duplicate images and an uneven image distribution, which hampers image-text alignment~\cite{laurenccon2024obelics}. This accounts for the inferior image-text alignment observed in OF compared to IDE in both internal and external experiments. OF tends to focus more on the text portion of in-context examples, neglecting the utilization of image information.

Additionally, dataset of OF contains sequences with an average of fewer tokens (less than 256)~\cite{awadalla2023openflamingo}, while OBELICS used by IDE features longer sequences (an average of 677 tokens)~\cite{laurenccon2024obelics}. This implies that IDE is better equipped to handle longer context windows during training, thereby leveraging the potential of many-shot inference.
These pre-training disparities lead to different behavioral performances of the two models in various settings. 
For example, When using many shots, OF is more likely to generate low-quality text and create hallucination. 
More behavioral differences will be discussed in detail in external and internal experiments analysis.

\subsubsection{Different capability levels of vision and language module of LMMs\label{visVslang}}
Current mainstream multimodal large models exhibit a systemic capability imbalance between their visual and language modules.
Taking OpenFlamingo-9B as an example, its language module employs the LLaMA/MPT architecture with 7 billion parameters, pre-trained on 1 trillion tokens of text corpora. In contrast, its visual module uses the CLIP ViT-L/14 encoder with only 428 million parameters and 400 million image-text pairs for pre-training. 
Similarly, IDEFICS-9B language module adopts LLaMA-7B, trained on 1.5 trillion tokens, while its visual encoder remains a CLIP-based model (OpenCLIP ViT-H/14) with 632 million parameters but the same limited pre-training data of 2 billion image-text pairs. 
This significant disparity in parameter scale and training data creates a structural gap in multimodal representation capacity. Empirical observations reveal that models tend to rely on linguistic priors over visual features\cite{li2024configure,lyu2022z}. During multimodal reasoning, the language module dominates information processing due to its stronger pattern recognition and semantic generation capabilities. 
This imbalance elevates visual entity recognition errors and exacerbates hallucination generation. 
In our experiments, we also identify the issue of differential contributions from the vision and language modules of LMMs during reasoning, which will be elaborated in Section \ref{external results}.

\begin{figure*}[htb]
\vspace{-5pt}
  \centering
  \includegraphics[width=1\linewidth, height=0.5\textheight, keepaspectratio]{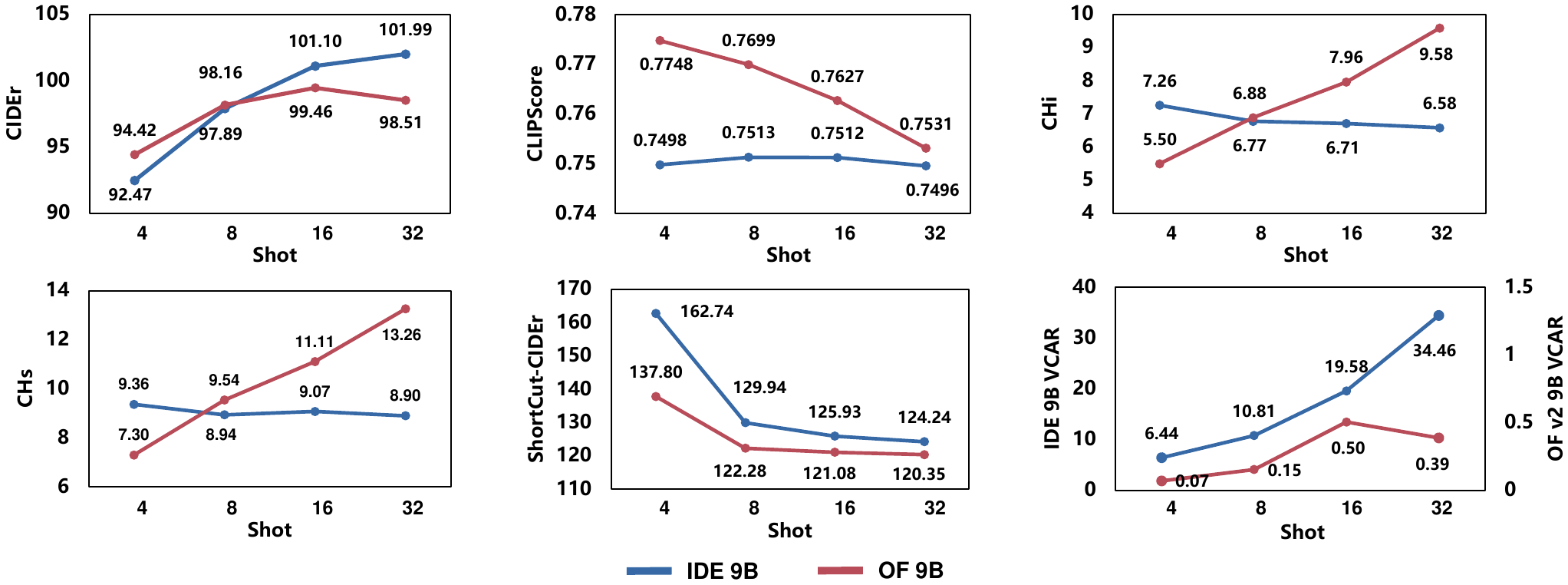}
  \vspace{-15pt}
  \caption{The experimental results from the shot perspective.}
  \vspace{-10pt}
  \label{fig:shot_results}
\end{figure*}

\subsubsection{Different Captioning Metrics\label{metrics}}
We use multiple metrics to evaluate the outputs. These include CIDEr~\cite{vedantam2015cider} and CLIPScore~\cite{hessel2021clipscore} for assessing output text quality; CHAIR~\cite{rohrbach2018chair} for measuring hallucination levels; Short-cut CIDEr, a metric we specifically propose, for gauging the proportion of demonstration text plagiarism; and the attention-based metric VCAR designed for determining the extent to which query visual information and demonstration text information are used.

Specifically, the quality of generated captions can be evaluated from two perspectives: linguistic and visual. The linguistic aspect involves grammatical correctness and the resemblance of linguistic patterns to human descriptions. The visual aspect assesses whether the caption accurately describes the main events in the image. Correspondingly, there are two common types of caption metrics. The first type, exemplified by CIDEr, focuses on the linguistic aspect by measuring the n-gram similarity between the generated and ground truth captions. However, these metrics may give inflated scores if the generated caption includes words or patterns frequently found in multiple ground truth captions.
The second type of metrics emphasize the visual aspect, with CLIPScore as a typical example. It calculates the cosine similarity between the embeddings of the caption and the image, obtained through the CLIP language and vision encoders. However, since CLIP~\cite{radford2021clip} is trained to assess whether an image and caption form a pair, it focuses on coarse-grained matching relationships. This makes it challenging to evaluate changes in caption-image alignment after minor variations.

In addition to the two common caption metrics, we employ CHAIR to specifically evaluate hallucinations. CHAIR measures the proportion of generated words that accurately represent the image, based on ground truth sentences and object segmentation. It includes two variants: CHAIRi (CHi), which measures the fraction of object instances hallucinated in each sentence, and CHAIRs (CHs), which assesses the fraction of sentences containing a hallucinated object.

Furthermore, we design short-cut CIDEr to measure the extent of text copy rate by LMMs from demonstrations. It is calculated by getting the CIDEr score between generated captions and ICE captions, directly showing the repetition rate of demonstration captions in generated text. To control variables, we only calculate the first four ICE captions in any shot setting.

VCAR, as stated in Section~\ref{internal}, is proposed to judge whether LMMs use more visual information from the query image or text information from in-context captions. We break down the attention weights of visual and text parts in the Flamingo~\cite{alayrac2022flamingo} architecture model in a special way. This is key to explaining hallucination and short-cut metrics since we believe that using less visual info from the query image and being overly guided by demonstration text are the main reasons for short-cuts.
It is worth noting that reduced VCAR levels or naturally low VCAR readings are frequently linked to heightened model hallucinations and short-cut inference in our experiment results. 
However, our analysis also indicates that these factors do not display a strictly inverse relationship. This implies that the root causes of hallucinations are multifaceted and can not be fully explained by the utilization of visual and text information alone.

\subsection{Holistic Conclusions from External Experiments\label{external results}}

Our exploration and conclusions of external ICEs configuration strategies is primarily divided into three dimensions. 
Firstly, we consider the number of in-context examples, or ``shots", which extends from few-shot scenarios with 4 shots to many-shot scenarios with 32 shots. 
Secondly, we examine caption assignment strategies, including FHL, MGC, and MHL, focusing on how different sources and qualities of captions paired with the same in-context image affect generation.
Lastly, we explore image retrieval strategies, specifically RS and SIIR, to understand the impact of the image similarity between ICE and query on the outcomes. 
Furthermore, the attention-based analysis metrics, particularly VCAR introduced in the earlier section, help clarify the behavioral characteristics exhibited in the external experiments.

In the forthcoming analysis phase, we will synthesize key findings from these three dimensions of external exploration and provide explanations for some phenomena through an analysis of internal attention maps and pre-training differences.
Given the combinatorial explosion of experimental configurations, we mitigate presentation complexity by reporting dimension-wise averaged results. For instance, when mentioning performance of OF under 4-shot setting, we present the mean across all combinations of two image retrieval methods and eleven caption assignment strategies. 

\begin{figure*}[htb]
\vspace{-10pt}
  \centering
  \includegraphics[width=1.0\linewidth]{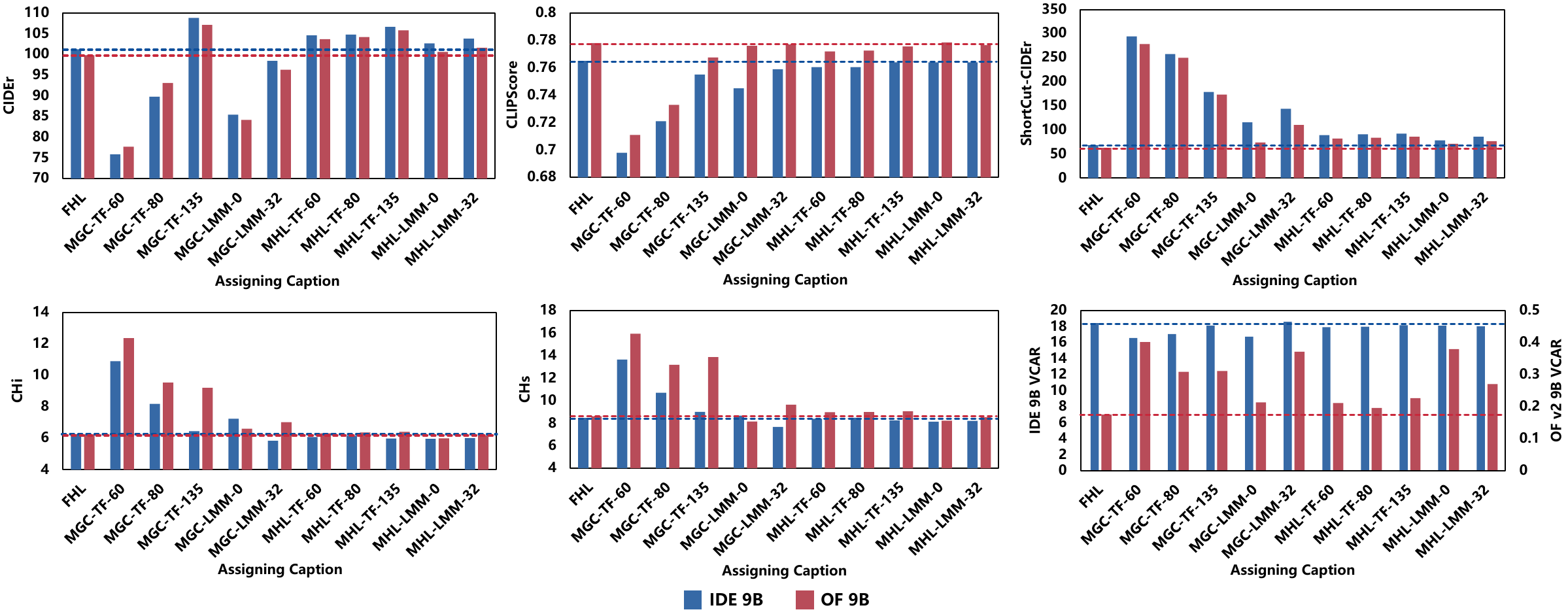}
  \vspace{-12pt}
  \caption{The experimental results from assigning caption perspective. The dashed line indicates the score of First Human-Labelled captions (FHL).}
  \vspace{-5pt}
  \label{fig:configuration_results}
\end{figure*}

\subsubsection{Number of Shots}
In an ideal scenario, intuitively, the more in-context examples provided, the better the in-context learning performance should be. However, in practical applications, models are influenced by their pre-training and are constrained by the context window, leading to unexpected performance in long-shot scenarios.

\begin{figure}[t]
 \vspace{0pt}
  \centering
  \includegraphics[width=1\linewidth]{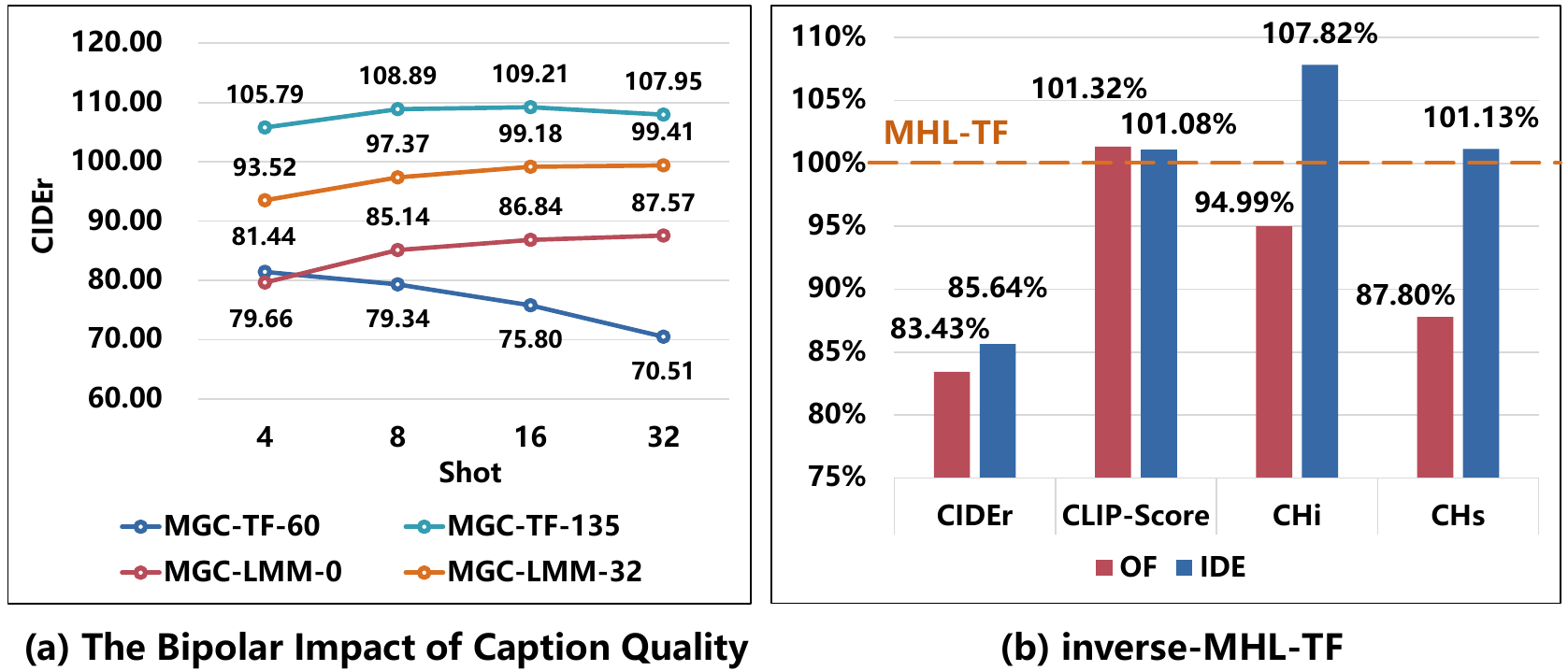}
  \vspace{-10pt}
  \caption{(a) demonstrates the bipolar impact of caption quality. High-quality ICE captions improve model performance, which gets better with more shots. Conversely, low-quality ICE captions worsen model performance, especially as the number of shots increases.
(b) presents the inverse MHL-TF experimental results. Due to the inconsistent units of different metrics, we use the average performance of MHL-TF as the baseline. The graph shows the percentage change of inverse MHL-TF relative to this baseline.
}
\vspace{-15pt}
  \label{fig:supplementary}
\end{figure}

As shown in Figure \ref{fig:shot_results}, increasing the number of shots generally improves CIDEr scores for both LMMs. 
This trend is accompanied by reduced hallucination in IDE, diminished short-cut phenomenon, and increased utilization of query visual information in both LMMs.
We believe it is due to the enhanced diversity of demonstrations, preventing the model from blindly copying ICE.
This suggests that increasing the number of shots indeed helps LMMs, but such benefits are limited. 
In scenarios involving low-quality captions (Figure \ref{fig:supplementary} (a), TF-60), increasing shots leads to deteriorating CIDEr and CLIPScores for both LMMs, indicating that excessive low-quality examples introduce noise and degrade visual-text alignment. 
Furthermore, while CIDEr scores improve with more shots, CLIPScores show diminishing returns and even decline in some cases. Notably, OF exhibits increased hallucination with higher shot counts, suggesting that additional linguistic context does not enhance visual understanding but merely provides superficial linguistic assistance.

\begin{tcolorbox}[colback=red!5!white,colframe=red!75!black]
 \textbf{Finding 1. Decoupled Modality Gains in a More-shot Setting.} 
 In multimodal in-context learning, increasing the number of shots improves linguistic coherence but degrades visual-text alignment and increases hallucination, indicating that shot augmentation primarily enhances linguistic pattern recognition rather than true multimodal reasoning.
 \end{tcolorbox}

\begin{figure*}[htb]
  \centering
  \includegraphics[width=1.0\linewidth]{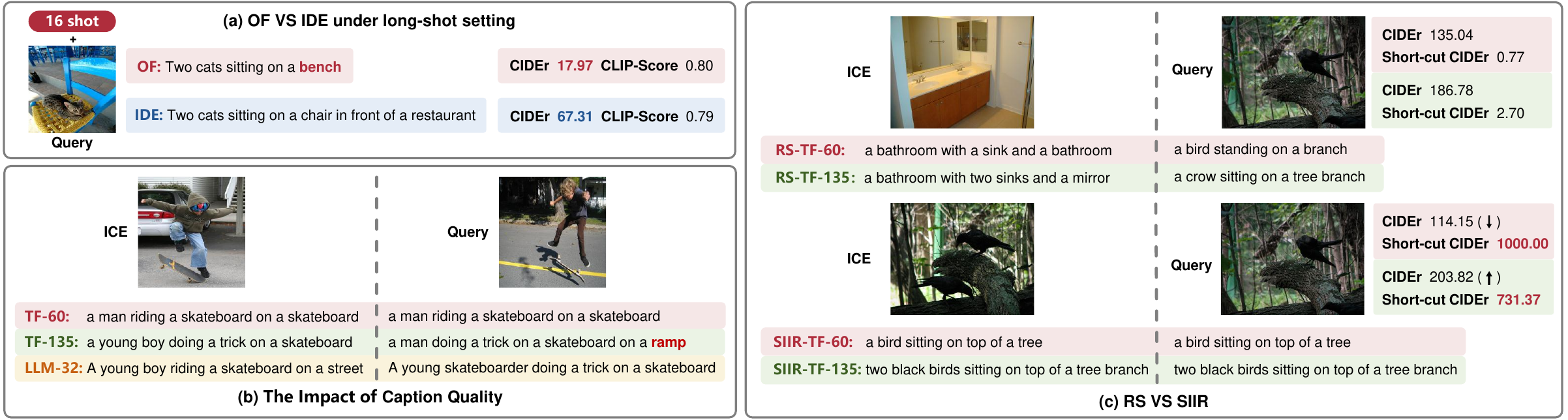}
  \vspace{-15pt}
  \caption{\textbf{Three visualization examples.} (a) Two captions generated by OF and IDE in 16-shot setting. The caption of OF has a low CIDEr score and contains hallucinations (``bench"), while that of IDE is more accurate.
(b) An example of OF under SIIR 32-shot setting. Using three quality levels of ICE captions (MGC-TF-60, MGC-TF-135, and MGC-LMM-32) as demonstrations. TF-60 misleads the model to generate low-quality sentences. TF-135 results in better sentence structures but with hallucinations (``ramp") and short-cuts. LMM-32 produces accurate, diverse-style sentences without obvious short-cuts. 
(c) An example of IDE under 4-shot setting. Model performance with RS and SIIR retrieval methods. Similar ICE images intensify the impact of ICE captions. Low-quality captions worsen performance, while high-quality ones enhance it. Meanwhile, SIIR significantly increases short-cut plagiarism.
}
\vspace{-10pt}
  \label{fig:visualization}
\end{figure*}

When comparing OF and IDE, we observe different performance trends as the number of shots increases. 
While both models show improvements in CIDEr scores with more shots, IDE demonstrates significantly greater gains than OF, particularly in long-shot scenarios where performance of OF plateaus or even declines, as shown in Figure~\ref{fig:shot_results}. Similarly, while CLIPScores decrease for both models as shots increase, IDE exhibits greater resilience against such declines. Notably, hallucination rate of OF rises sharply with additional shots, whereas IDE remains stable and even shows a slight reduction in hallucination under long-shot conditions.In Figure \ref{fig:visualization} (a), OF generates low-CIDEr captions with evident hallucinations (``bench") under 16-shot ICEs, while IDE produces more accurate outputs.

These findings suggest that while increasing shots generally aids model performance, OF struggles with long contexts, likely due to its pre-training data limitations as described in Section~\ref{OFvsIDE}. Specifically, pre-training dataset of OF contains significantly shorter sequences (under 256 tokens) compared to OBELICS (averaging 677 tokens), limiting the ability of OF to handle extended context windows. Additionally, VCAR values of OF remain below 1, indicating minimal utilization of visual information, whereas VCAR values of IDE exceed tens, reflecting extensive visual reliance. This disparity aligns with differences in pre-training data quality: MMC4 contains numerous duplicate images and uneven image distributions, impairing effective image-text alignment and forcing OF to over-rely on linguistic components during inference.

These results highlight that even architecturally identical models can exhibit markedly different behaviors due to biases in their pre-training corpora. Notably, the reliance of OF on linguistic patterns, compensating for its weaker visual capabilities, exacerbates hallucination issues, whereas stronger visual-text integration of IDE enables better performance stability in many-shot scenarios.

\begin{tcolorbox}[colback=red!5!white,colframe=red!75!black]
 \textbf{Finding 2. Pre-training Corpus Bias Induces Multimodal Disparity. } 
 Architecturally identical LMMs exhibit distinct behaviors during testing due to biases in pre-training corpora. Specifically, models trained on shorter sequences and lower-quality image-text pairs struggle with long-context reasoning and visual-text alignment, leading to increased hallucination and performance instability.
 \end{tcolorbox}


\subsubsection{Demonstration Caption Quality}
In our experiments on caption assignment strategies, we have found that the quality of ICE captions significantly impacts the performance of LMMs. To evaluate this effect, we have trained a small Transformer (TF) model and used CIDEr scores of its generated captions as a quality metric, defining three quality levels: TF-60 (low), TF-80 (medium), and TF-135 (high). Additionally, we also use captions generated by the LMM itself under 0-shot (LMM-0, low-quality) and 32-shot (LMM-32, high-quality) settings. 
Results are shown in Figure \ref{fig:configuration_results}.



Our observations reveal a complex relationship between ICE caption quality and model performance. 
First, ICE quality has a bifurcated impact on LMM behavior. As demonstrated in Figure \ref{fig:supplementary} (a), low-quality captions degrade performance as the number of shots increases, while high-quality captions enhance results even in long-shot scenarios. Notably, high-quality ICE partially compensates for limitations of OF in handling long contexts, highlighting the potential benefits of improving ICE quality.

Second, we observe significant trade-offs across evaluation metrics. While certain ICEs configurations improve CIDEr scores (e.g., MGC-TF-135), these gains often come at the cost of reduced CLIPScores and increased hallucination and short-cut inference. For instance, in Figure~\ref{fig:configuration_results}, the MGC-TF-135 setting achieves the highest CIDEr scores but performs poorly in visual-text alignment (CLIPScore) and exhibits substantial hallucination and shortcut phenomena. This suggests that improvements in linguistic coherence do not necessarily translate to better visual-text alignment and multimodal reasoning.
Furthermore, our experiments indicate limitations in multi-round generation strategies. While some configurations (MGC) improve CIDEr scores, they fail to address visual fidelity or hallucination issues, suggesting diminishing returns for such approaches. 
This sensitivity to text quality likely stems from stronger reliance of LMMs on linguistic processing over visual reasoning, as detailed in Section \ref{visVslang}.



\begin{tcolorbox}[colback=red!5!white,colframe=red!75!black]
 \textbf{Finding 3. ICE Quality Sensitivity Triggers Multimetric Conflicts. } 
 LMMs exhibit heightened sensitivity to ICE caption quality due to their linguistic dominance in reasoning. The influence of ICE quality on model performance is multifaceted, often resulting in trade-offs across metrics. Improvements in linguistic coherence may come at the cost of visual-text alignment and increased hallucination, underscoring the limitations of relying solely on ICE quality enhancements for performance optimization.
 \end{tcolorbox}

Our experiments also reveal that the linguistic patterns of ICE significantly influence LMM reasoning, using ICE captions with certain metric preferences can lead the model to generate sentences with inflated scores on that metric. 
To explore this, we have designed an inverse MHL-TF experiment using ground truth captions least similar to TF-generated ones, ensuring CIDEr-unbiased and human-labeled ICE captions. Results in Figure \ref{fig:supplementary} (b) show decreased CIDEr scores but increased CLIPScores and reduced hallucination for OF. This indicates that LMMs, due to their auto-regressive pre-training, focus on linguistic structures, potentially neglecting semantics. Thus, diverse ICE captions can enhance visual accuracy and prevent metric inflation.

Conversely, in MGC-LMM and MHL-LMM experiments where LMM-generated captions served as ICE captions, we observe better CLIPScores, fewer hallucinations and short-cuts compared to TF-generated captions. As shown in Figure \ref{fig:visualization} (b), while TF-60 results in poor-quality generation and causes short-cuts, TF-135 introduces hallucinations (``ramp"), and LMM-32 produces high-quality sentences without significant issues. 
This suggests that LMM-generated captions, aligning with the inherent style of model, avoid disruptions from external knowledge, leading to more stable reasoning.
Finally, upon comprehensive observation as we previously mention, the MHL strategy have emerged as optimal caption assignment strategy. By using LMM or small model-generated captions as anchors and selecting ground truth captions with similar patterns, MHL combines machine strengths with human accuracy, ensuring stable improvements over the baseline.


\begin{tcolorbox}[colback=red!5!white,colframe=red!75!black]
 \textbf{Finding 4. ICE Linguistic Style Imitation Undermines Visual Fidelity.} 
 The linguistic patterns of ICE captions significantly impact LMM reasoning, causing models to imitate sentence styles at the expense of visual accuracy. Using diverse, model-consistent captions can alleviate this problem.
 \end{tcolorbox}




\begin{figure*}[htb]
\vspace{-10pt}
  \centering
  \includegraphics[width=1.0\linewidth, height=0.7\textheight, keepaspectratio]{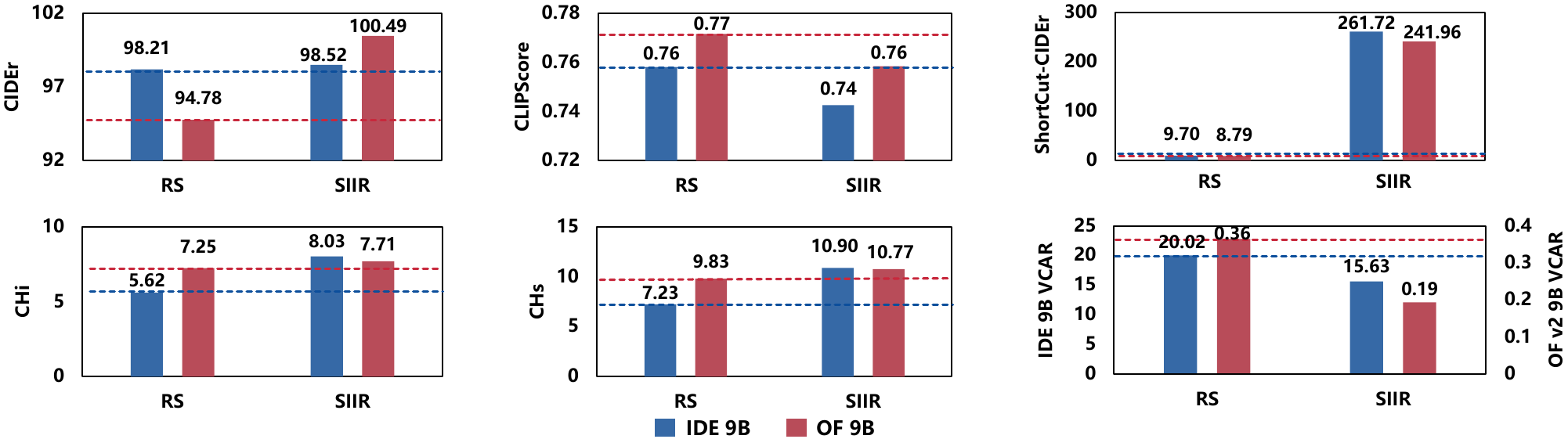}
  \vspace{-10pt}
  \caption{The experimental results from the image retrieval perspective. The dashed line indicates the score of Random Sampling (RS).
}
\vspace{-10pt}
  \label{fig:retrieval_results}
\end{figure*}

\subsubsection{Image Similarity}
Studies in NLP have indicated that similarity-based ICE retrieval methods significantly enhance model performance. However, unlike single modal, multimodal research is more complex. our comprehensive experiments reveal that the improvements brought by similarity-based retrieval are not straightforward in multimodal contexts.

As shown in Figure~\ref{fig:retrieval_results}, comparing RS and SIIR reveals that using visually similar ICE images increases CIDEr scores for both LMMs. However, CLIPScores show no consistent improvement and even decline. Additionally, SIIR introduces more hallucinations and short-cuts compared to RS, and VCAR values decrease significantly, indicating reduced attention to query images.
These findings suggest that similarity-based retrieval primarily benefits linguistic coherence (CIDEr) but not semantic accuracy (CLIPScore). The improvements may stem from short-cut reasoning, where models copy sentences from ICE rather than genuinely understanding visual content. 
For example, in figure~\ref{fig:visualization} (c), when using SIIR setting, the generated text is strikingly similar to the ICE caption, almost identical, and its short-cut CIDEr is nearly 1000 times that of the RS setting.
This leads to higher CIDEr scores but worsens visual-text alignment and increases hallucination.

\begin{tcolorbox}[colback=red!5!white,colframe=red!75!black]
 \textbf{Finding 5. Similarity-Based Retrieval Induces Shortcut-Driven CIDEr Inflation. } 
 While using visually similar ICE images increases CIDEr scores, this is essentially due to short-cut phenomenon where captions from ICE are copied. This undermines the utilization and accuracy of visual information and exacerbates hallucination.
 \end{tcolorbox}


In previous experiments, we have observed that when the quality of ICE captions is low, the quality of the captions generated by LMMs also diminishes.
Our experiments demonstrate that the similarity of ICE images can significantly amplify the impact of ICE captions on model reasoning. Specifically, under the SIIR setting, when ICE captions are of low quality, the generation quality further diminishes, as shown in Figure \ref{fig:visualization} (c) (TF-60). Conversely, high-quality ICE captions enhance the CIDEr scores of generated captions (TF-135). This suggests that similar ICE images intensify the influence of ICE captions on reasoning.
We speculate that this phenomenon originates from the cross-attention mechanism of Flamingo framework. As shown in Figure \ref{fig:flamingo_xatt}, visual information is integrated into captions through cross-attention. Here, captions act as \textbf{Q}ueries, while image features serve as \textbf{K}eys and \textbf{V}alues. The output is essentially a linear combination of visual features (\textbf{V}alues). When ICEs and the query image are highly similar, the cross-attention outputs become nearly identical, which can hinder accurate comprehension and reasoning.


Furthermore, IDE exhibits heightened sensitivity to similar images compared to OF. SIIR causes a more significant decline in CLIPScores and introduces more hallucinations and short-cuts in IDE. 
This heightened sensitivity likely stems from differences in pre-training datasets. As discussed in Section~\ref{OFvsIDE} regarding the MMC4 dataset, the pre-training dataset for IDE contains higher-quality image-text pairs. Consequently, IDE achieves better modality alignment and stronger visual understanding capabilities.
These findings highlight that models with stronger visual understanding capabilities are more susceptible to the amplifying effect of similar ICE images. This effect, rooted in cross-attention mechanisms, causes models to over-rely on ICE text when images are visually similar, thereby magnifying both the benefits and drawbacks of ICE captions.

\begin{tcolorbox}[colback=red!5!white,colframe=red!75!black]
 \textbf{Finding 6. Similarity Retrieval Amplifies ICE Caption Influence.} 
 The similarity of ICE images can amplify the impact of ICE captions on reasoning due to cross-attention multimodal architecture. Models with stronger visual understanding capabilities, such as IDEFICS, are more susceptible to this influence.
 \end{tcolorbox}

 \subsection{Significant Results in Internal Analyses\label{internal results}}

In Section~\ref{internal}, we identify and examine three key internal characteristics of LMMs, designing specialized attention-based metrics for these purposes. ACAR and IEAR directly assess the degree of Anchor Token and Emergent Attention Window patterns, while VCAR calculates the ratio of query visual information to demonstration text information used during inference, potentially explaining short-cut reasoning and hallucinations.
Given that short-cut inference and VCAR are thoroughly discussed in Section~\ref{external results} with external experimental results, this section focuses solely on the experimental results related to Anchor Token and Emergent Attention Window.

\subsubsection{Anchor Token}

\begin{figure}[t]
\vspace{0pt}
  \centering
  \includegraphics[width=1.0\linewidth]{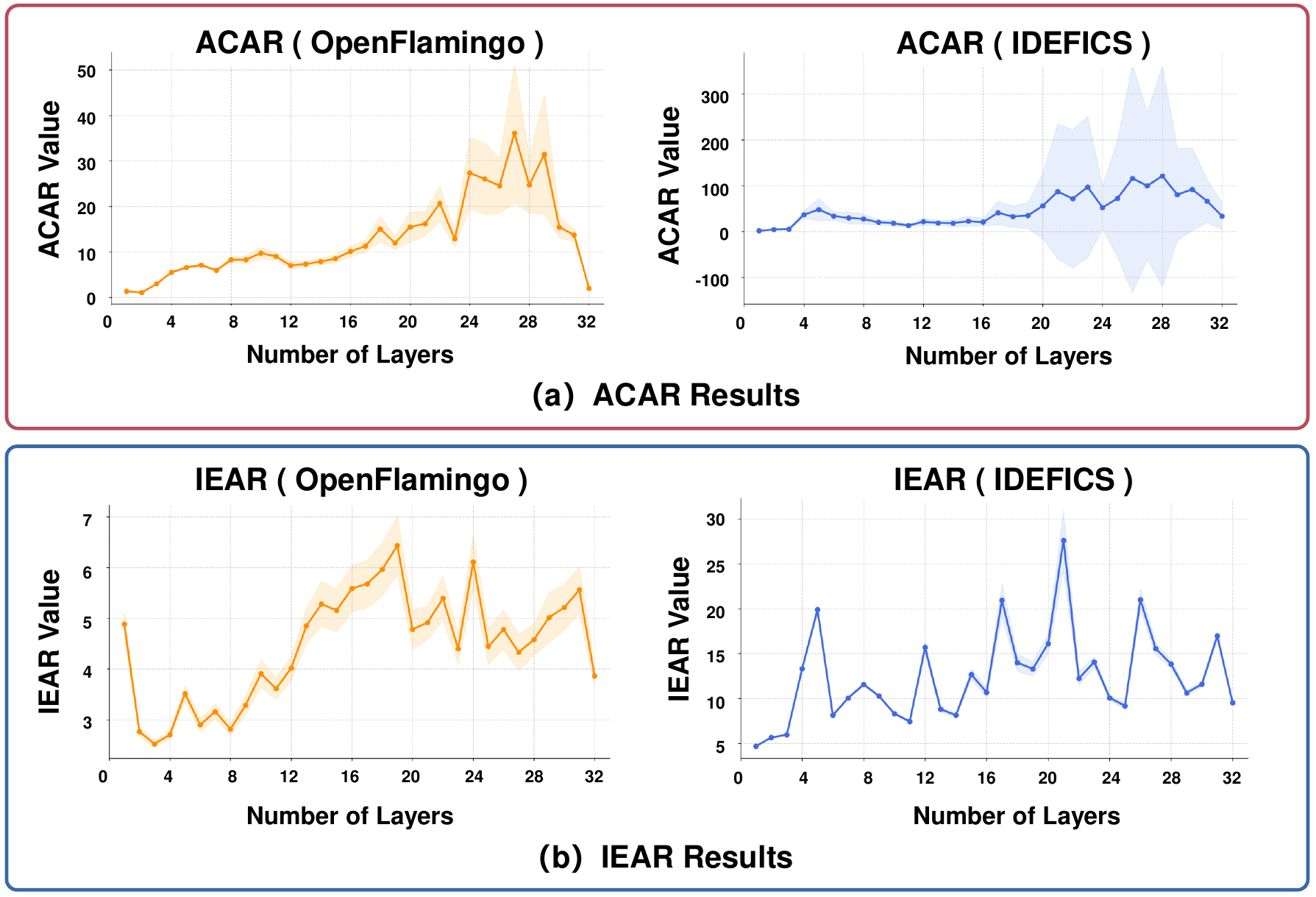}
  \vspace{-15pt}
  \caption{The average results of ACAR and IEAR for two models at each layer.
}
\vspace{-12pt}
  \label{fig:ar_results}
\end{figure}

To quantify the anchor token aggregation observed in attention maps, we developed the ACAR metric based on attention weights as detailed in Section~\ref{internal}. Figure~\ref{fig:ar_results} (a) shows the mean ACAR curve across layers, revealing that attention values on anchor tokens are significantly larger than those on other context tokens. This confirms the attention aggregation effect on anchor tokens. 
Despite the consistent higher attention values on anchor tokens compared to context tokens, ACAR reaches particularly high values in the middle and later layers. The anchor phenomenon is more prominent in deeper layers than in shallower ones, indicating that there are more contextual interactions in earlier layers, while certain anchor tokens become carriers of attention information in deeper layers, prioritized during the forward pass to relay information.
Meanwhile, IDE exhibits a more pronounced anchor aggregation phenomenon than OF. 

\begin{table}[t]
    \centering
    \vspace{0pt}
    \caption{Selective Masking Results. ``Anchor-centric masking" keeps only anchor tokens and masks others. ``Context-centric masking" restricts each ICE to attend only to its own tokens, blocking interaction with other ICEs. Numbers in parentheses indicate the layers where masking is applied.}
    \scalebox{0.75}{
    \begin{tabular}{lllll}
    \hline
        Shot & 4 & 8 & 16 & 32  \\ \hline
        Baseline & 97.3 & 101.7 & 103.8 & 104.8  \\
        Anchor-Centric Masking(10-30) & 74.1(-23.8\%) & 77.6(-23.7\%) & 80.9(-22.1\%) & 82.8(-21.0\%)  \\ 
        Anchor-Centric Masking(15-30) & 94.8(-2.6\%) & 98.8(-2.8\%) & 101.6(-2.1\%) & 102.7(-2.0\%) \\ 
        Context-Centric Masking(10-30) & 97.1(-0.2\%) & 101.4(-0.2\%) & 102.9(-0.9\%) & 105.1(+0.3\%) \\ \hline
    \end{tabular}
     }
    \vspace{-5pt}
    \label{selective-masking}
\end{table}

\begin{table}[t]
    \centering
    \caption{Model Pruning Results. All results are tested on IDEFICS under RS-4 shot settings (1000 samples, batch size = 1). Numbers after ``Fast" indicate the starting layer for pruning. ``w/o recover" means pruned tokens are not recovered in the prediction layer. ``Time" is the average generation time per sample, and ``KV cache" specifies the memory usage of KV cache for the current method.}
    \begin{tabular}{lllll}
    \hline
        sample=1000 & Shot & CIDEr & Time & KV cache \\ \hline
        Baseline & 4 & 100.1 & 1.28 s/it & 78.64M  \\ 
        Fast\_10(w/o recover) & 4 & 55.5 & \textbf{1.18 s/it} & \textbf{41.15M}(-47.7\%) \\ 
        Fast\_10 & 4 & 61.1 & 1.21 s/it & 42.86M(-45.5\%) \\ 
        Fast\_15 & 4 & 95.4 & 1.19 s/it & 51.38M(-34.7\%) \\ 
        Fast\_20 & 4 & \textbf{100.5} & 1.20 s/it & 59.89M(-23.8\%) \\ \hline
        Baseline & 32 & \textbf{108.1} & 3.71s/it & 548.41M  \\ 
        Fast\_10(w/o recover) & 32 & 59.72 & \textbf{3.18s/it} & \textbf{248.51M}(-54.7\%) \\ 
        Fast\_10 & 32 & 69.95 & 3.20s/it & 262.14M(-52.2\%) \\ 
        Fast\_15 & 32 & 101.3 & 3.21s/it & 330.30M(-39.8\%)\\ 
        Fast\_20 & 32 & 108.0 & 3.27s/it & 398.45M(-27.3\%)\\   
        \hline
    \end{tabular}
    
    \label{Anchor: model pruning}
\end{table}

\begin{figure}[t]
  \centering
  \includegraphics[width=1.0\linewidth]{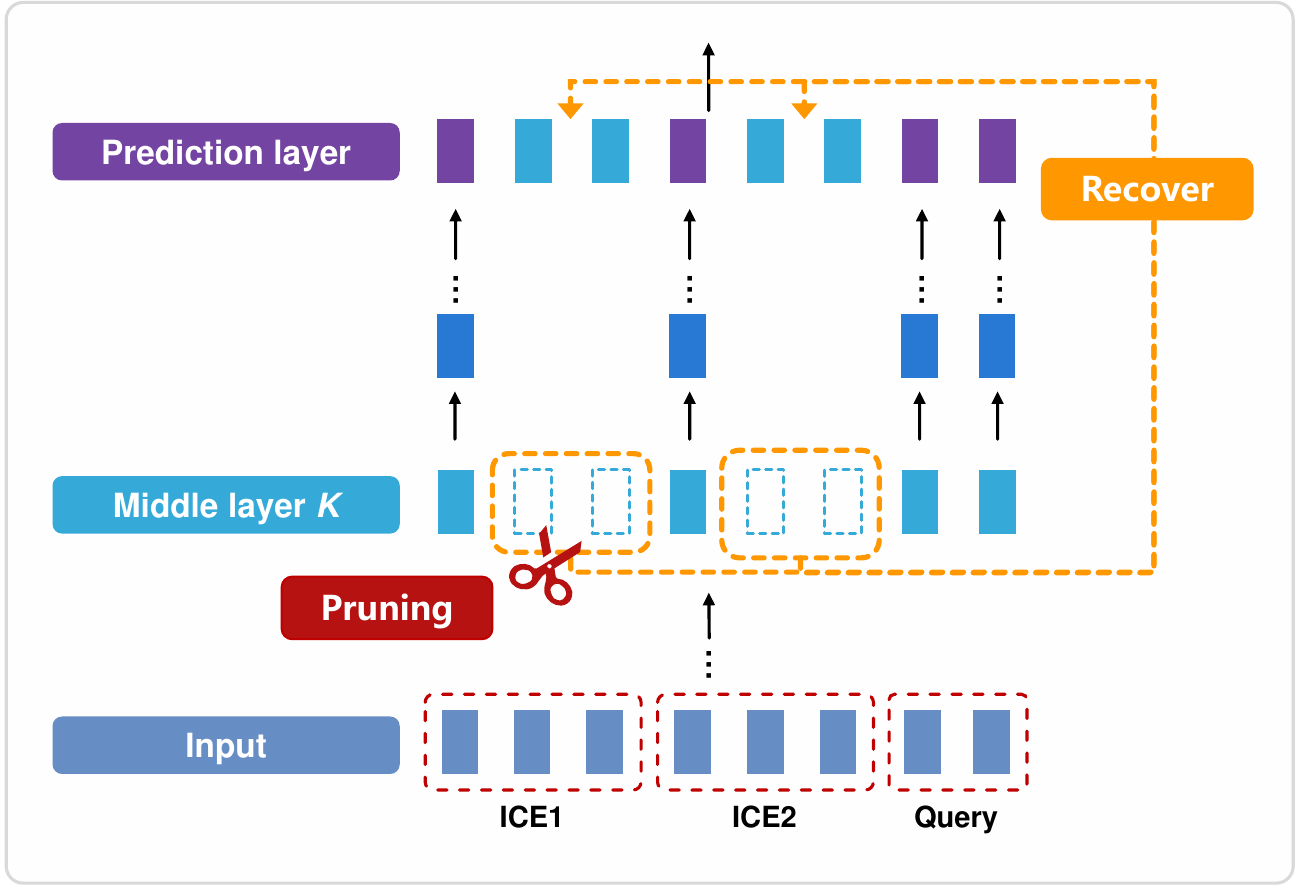}
  \vspace{-10pt}
  \caption{\textbf{Model pruning.} To accelerate inference, non-anchor tokens in ICE are pruned at middle layer \(\mathcal{K}\). All tokens are then restored at the prediction layer using the hidden states of pruned layer to maintain performance.
}
\vspace{-10pt}
  \label{fig:model pruning}
\end{figure}

Inspired by this pattern and previous work on LLMs~\cite{ge2023llmtokenprune,ma2023llmtokenprune,syed2023llmtokenprune,xiao2024LLMtokenprune,frantar2023llmtokenprune}, we explore using anchor tokens for inference acceleration in LMM ICL. 
We first conduct selective masking experiments by obscuring all tokens in the middle layers of the decoder except for the anchor and query tokens. Results in Table~\ref{selective-masking} indicate that this masking does not significantly degrade performance (Anchor-Centric Masking). 
We then implement an unstructured model pruning strategy. As shown in Figure~\ref{fig:model pruning}, we retain only the anchor and prediction tokens from the $k$-th layer and prune other context tokens. To ensure reasoning effectiveness, we recover the hidden states of all pruned tokens before the prediction layer using their hidden states from the $k$-th layer. This approach reduces the input sequence length by over 70\%. For example, in a 32-shot setting, the average sequence length is 1046 tokens, of which only 192 are left after pruning. This reduces computational load and cache usage by 50\% without significantly affecting performance, as shown in Table~\ref{Anchor: model pruning}. 
However, despite the significant reduction in sequence length and KV cache occupancy, the acceleration effect is not as pronounced as expected (\eg, 3.71s/it→3.20s/it), and the specific reasons will be analyzed in the next section.

Besides, we examine whether different external configuration strategies would affect the emergence of anchor tokens and the aggregation patterns. 
Observations show that anchor tokens are not affected by changing the ICEs and the method of inference accelerating consistently work for by external configuration strategies.

\subsubsection{Emergent Attention Window}

As noted in Section~\ref{internal}, each ICE token in demonstrations predominantly interacts within its own context, rarely engaging with other contexts. This creates an Emergent Attention Window pattern in attention maps. To quantify this, we have developed the IEAR metric.
Figure~\ref{fig:ar_results} (b) shows that both models exhibit stronger attention interactions within ICE than between ICEs, indicating the Emergent Attention Window effect. This relationship reaches very high values in the middle and later layers, particularly for IDEFICS (reaching 20x).

Some studies in NLP observe the phenomenon of Emergent Attention Window~\cite{zhu2024anchor&window} and explore to use attention window to achieve many-shot ICL compression~\cite{hao2022windownot,ratner2022windownot}. 
Specifically, they replace the global window over all the input tokens with a local window that only calculates the self-attention over the tokens in each ICE. Then if an input sequence contains $M$ ICEs and each ICE has $N$ tokens in average and assume that self-attention costs $O(N^2)$ computation complexity over $N$ tokens, replacing global window with local window will reduce the complexity from $O((MN)^2)$ to $O(MN^2)$.

Since Emergent Attention Window also appears in LMMs, we think whether this phenomena has potential to achieve acceleration. To validate this, we also use the local window masking and find that the performance does not suffer significant degradation as table\ref{selective-masking} (Context-Centric Masking). 


However, despite its theoretical potential, the acceleration effect is currently limited, facing the same problem as the drop token method we have proposed before.
We hypothesize two main reasons.
First, KV-cache allows attention for inputs to be calculated only once during inference. 
Second, current open-sourced LMMs do not handle long input sequences as effectively as LLMs. For instance, while a 32-shot setting nearly reaches the upper limit of our tested models, its length is only about 1K tokens, which makes the benefits of pruning acceleration strategies less significant. Consequently, dropping context tokens or reducing attention calculation complexity in multimodal ICL does not provide the same speed advantages as in many-shot scenarios for LLMs (\eg, 1000 shots\cite{bertsch2024manyshot}). Nevertheless, anchor-centric pruning or using local windows still shows potential for ICL acceleration when future LMMs become more capable of handling many-shot scenarios.

\section{Conclusion}

In this paper, we have carried out a comprehensive study on multimodal in-context learning for Large Multimodal Models for image captioning. Through systematic external and internal investigations, we have explored the impact of In-Context Example configurations on model performance and analyzed the internal attention mechanisms of LMMs. 
Externally, we have utilized multiple metrics to evaluate model performance, revealing key insights into how different ICEs configurations affect LMMs. 
Internally, we have identified typical attention patterns and developed attention-based metrics to quantify model behaviors. 
Our work demonstrates that both external and internal analysis perspectives are crucial for understanding and optimizing multimodal ICL in LMMs. The findings and proposed metrics offer valuable guidance for future research and development in LMMs, potentially applicable to broader research areas beyond image captioning.

\bibliographystyle{IEEEtran}
\bibliography{IEEEabrv}












\newpage

\section{Biography Section}
 

\vspace{-40pt}
\begin{IEEEbiography}[{\includegraphics[width=1in,height=1.25in,clip,keepaspectratio]{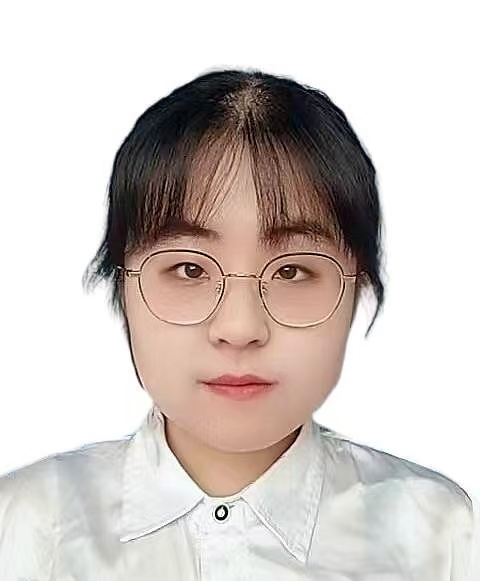}}]{Li Li}
received the B.Eng. degree from Bell Honors School, Nanjing University of Posts and Telecommunications in 2023. She is currently pursuing the M.Eng. degree at the School of Computer Science and Engineering, Southeast University, affiliated with the Key Lab of New Generation Artificial Intelligence Technology \& Its Interdisciplinary Applications (Ministry of Education). Her research interests mainly include large multimodal models, in-context learning, and multimodal understanding.
\end{IEEEbiography}
\vspace{-25pt}
\begin{IEEEbiography}[{\includegraphics[width=1in,height=1.25in,clip,keepaspectratio]{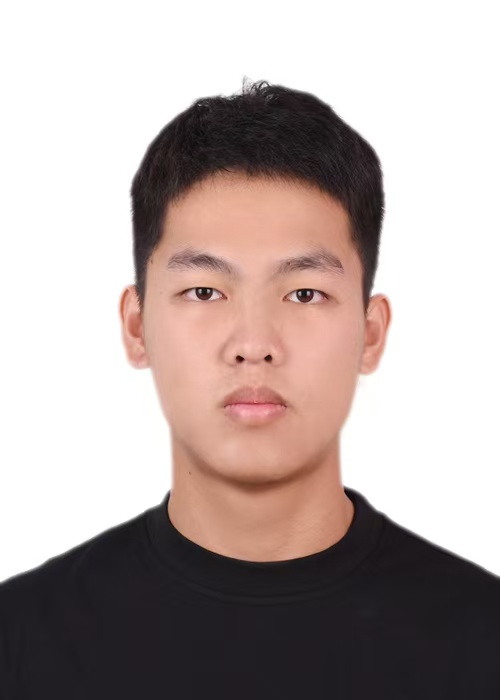}}]{Yongliang Wu}
is currently pursuing the M.Eng. degree at the School of Computer Science and Engineering, Southeast University, affiliated with the Key Lab of New Generation Artificial Intelligence Technology \& Its Interdisciplinary Applications (Ministry of Education). He received the B.S. degree in Artificial Intelligence from Southeast University in 2023. His research interests mainly include video understanding, in-context learning, machine unlearning, and image editing.
\end{IEEEbiography}
\vspace{-25pt}
\begin{IEEEbiography}[{\includegraphics[width=1in,height=1.25in,clip,keepaspectratio]{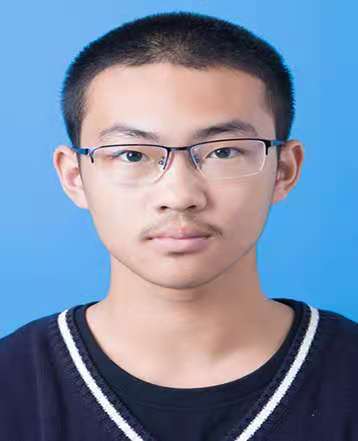}}]{Jingze Zhu}
received the B.Eng. degree from the School of Computer Science and Engineering,, Southeast University in 2024. He is currently pursuing the M.Eng. degree at the School of Computer Science and Engineering, Southeast University, affiliated with the Key Lab of New Generation Artificial Intelligence Technology \& Its Interdisciplinary Applications (Ministry of Education). His research interests mainly include large language model, vision language model and in-context learning.
\end{IEEEbiography}
\vspace{-25pt}
\begin{IEEEbiography}[{\includegraphics[width=1in,height=1.25in,clip,keepaspectratio]{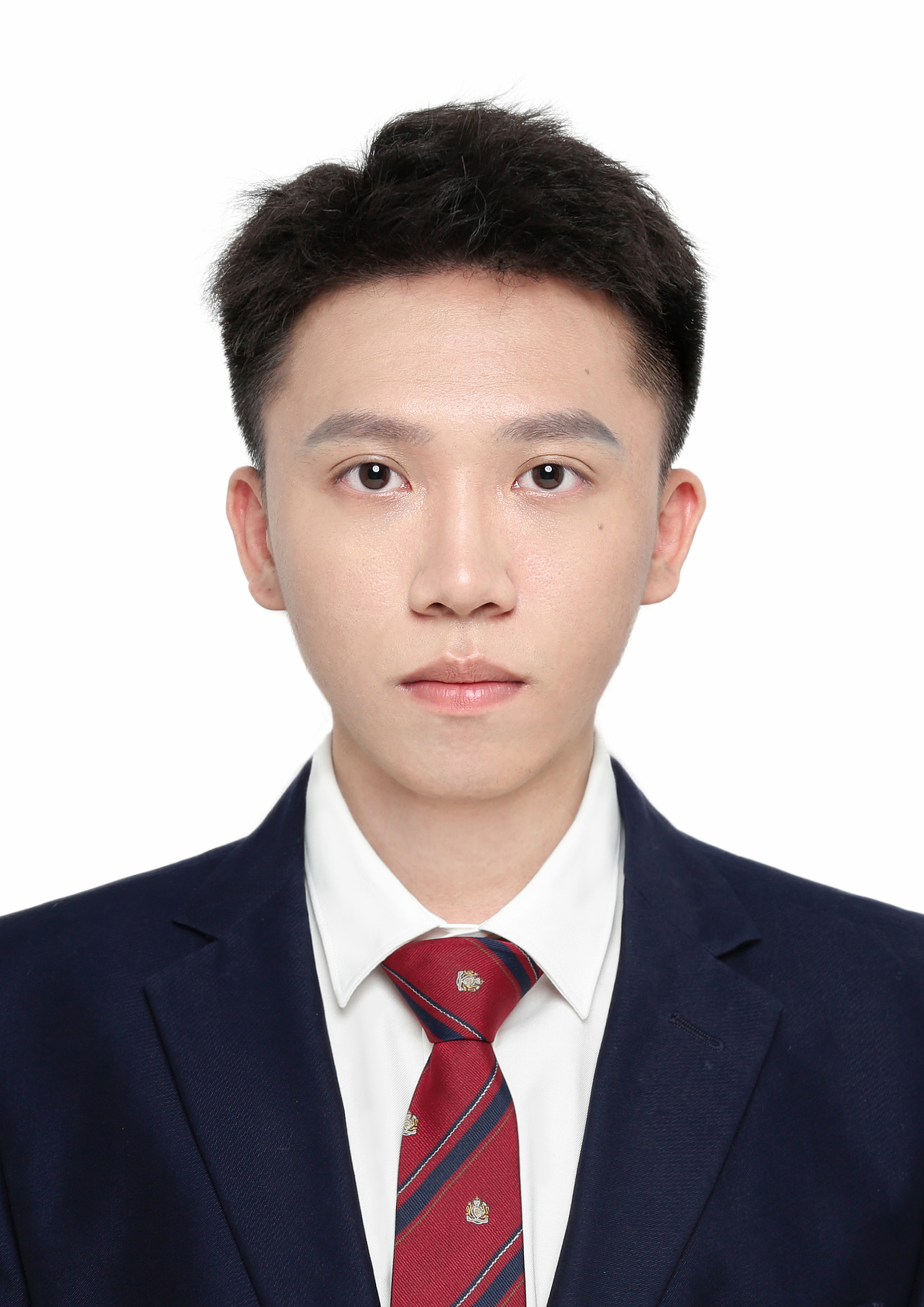}}]{Jiawei Peng} 
received the B.Eng. degree from Chien-Shiung Wu College, Southeast University in 2023. He is currently pursuing the M.Eng. degree at the School of Computer Science and Engineering, Southeast University, affiliated with the Key Lab of New Generation Artificial Intelligence Technology \& Its Interdisciplinary Applications (Ministry of Education). His research interests mainly include large multimodal models, in-context learning, and multimodal understanding.
\end{IEEEbiography}
\vspace{-25pt}
\begin{IEEEbiography}[{\includegraphics[width=1in,height=1.25in,clip,keepaspectratio]{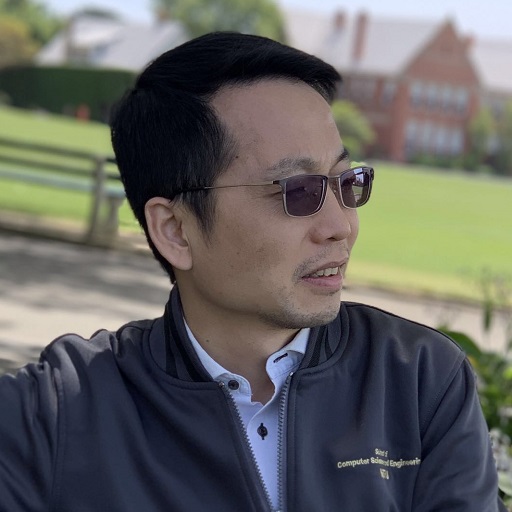}}]{Jianfei Cai}
received the PhD degree from the University of Missouri-Columbia. He is currently a professor with the Faculty of IT, Monash University, where he has served as the inaugural head of the Data Science \& AI Department. Before that, he was head of the Visual and Interactive Computing Division and head of the Computer Communications Division at Nanyang Technological University (NTU). His major research interests include computer vision, deep learning, and multimedia.  He is a co-recipient of paper awards in ACCV, ICCM, IEEE ICIP, and MMSP, and the winner of Monash FIT’s Dean’s Researcher of the Year Award. He serves or has served as an associate editor for IEEE Transactions on Pattern Analysis and Machine Intelligence, International Journal of Computer Vision, IEEE Transactions on Image Processing, IEEE Transactions on Multimedia, and IEEE Transactions on Circuits and Systems for Video Technology as well as serving as senior/area chair for CVPR, ICCV, ECCV, IJCAI, ACM Multimedia. He was the chair of IEEE CAS VSPC-TC during 2016-2018. He was the leading TPC chair for IEEE ICME 2012 and the leading general chair for ACM Multimedia 2024. He is a Fellow of IEEE.
\end{IEEEbiography}
\vspace{-25pt}
\begin{IEEEbiography}[{\includegraphics[width=1in,height=1.25in,clip,keepaspectratio]{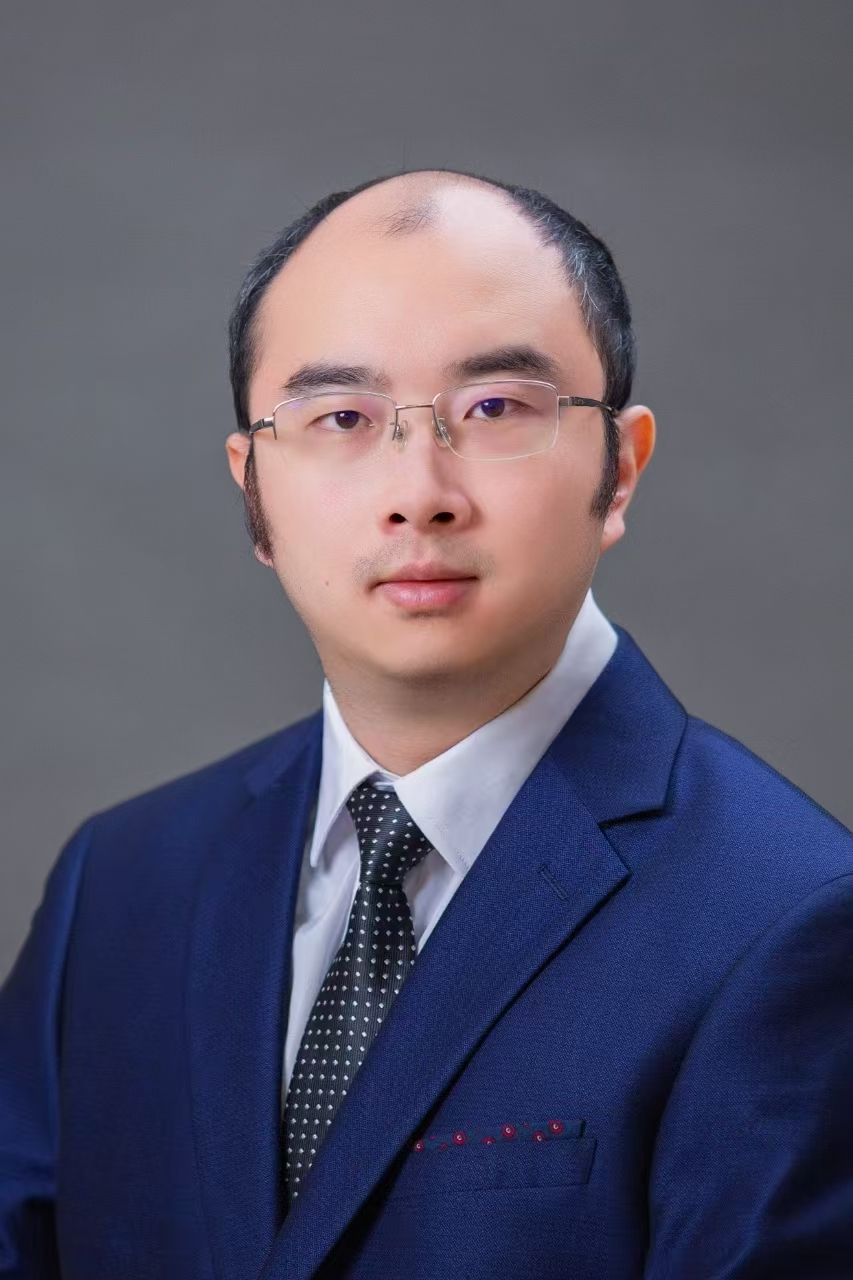}}]{Xu Yang}
is an associate professor at Southeast University. He received his Ph.D. from Nanyang Technological University, under the supervision of Prof. Jianfei Cai and Prof. Hanwang Zhang. His research interests include multi-modal in-context learning, vision-language models, and image captioning. He serves as a Senior Program Committee Member for IJCAI 2025, and an Area Chair for ACM Multimedia (ACMMM) 2024 and 2025. He was also named among the World’s Top 2\% Scientists in 2024.
\end{IEEEbiography}

\vspace{11pt}


\vfill

\end{document}